\documentclass[a4paper,twoside]{article}

\usepackage{epsfig}
\usepackage{subcaption}
\usepackage{calc}
\usepackage{amssymb}
\usepackage{amstext}
\usepackage{amsmath}
\usepackage{amsthm}
\usepackage{multicol}
\usepackage{pslatex}
\usepackage{apalike}
\usepackage{algorithm2e}
\usepackage[bottom]{footmisc}
\usepackage{url}
 \usepackage{multirow}
 \usepackage{tabularx}
 \usepackage{booktabs}
 \usepackage{float}
 \usepackage{verbatim}

\usepackage{fancyvrb}
\usepackage{xcolor}

\usepackage{SCITEPRESS}     

\begin{document}

\title{Informative Rays Selection for Few-Shot Neural Radiance Fields}

\author{\authorname{Marco Orsingher\sup{1,3}, Anthony Dell'Eva\sup{2,3}, Paolo Zani\sup{3}, Paolo Medici\sup{3} and Massimo Bertozzi\sup{1}}
\affiliation{\sup{1}University of Parma, Italy $\quad$ \sup{2}University of Bologna, Italy $\quad$ \sup{3}VisLab, an Ambarella Inc. company}
\email{morsingher@ambarella.com}
}



\abstract{Neural Radiance Fields (NeRF) have recently emerged as a powerful method for image-based 3D reconstruction, but the lengthy per-scene optimization limits their practical usage, especially in resource-constrained settings. Existing approaches solve this issue by reducing the number of input views and regularizing the learned volumetric representation with either complex losses or additional inputs from other modalities. In this paper, we present \textit{KeyNeRF}, a simple yet effective method for training NeRF in few-shot scenarios by focusing on key informative rays. Such rays are first selected at camera level by a view selection algorithm that promotes baseline diversity while guaranteeing scene coverage, then at pixel level by sampling from a probability distribution based on local image entropy. Our approach performs favorably against state-of-the-art methods, while requiring minimal changes to existing NeRF codebases.}

\onecolumn \maketitle \normalsize \setcounter{footnote}{0} \vfill

\section{\uppercase{Introduction}}
\label{sec:introduction}

3D reconstruction and novel view synthesis from a set of calibrated images is a longstanding challenge in computer vision, with applications in robotics \cite{3drob1,3drob2}, virtual reality \cite{3daug1,3daug2} and autonomous driving \cite{iciap,iv}. Recently, Neural Radiance Fields (NeRF) \cite{nerf} have been introduced to model 3D scenes with a small neural network that can be queried with any point in space to produce the corresponding density and view-dependent color. In order to enable end-to-end training from images, differentiable rendering is used to integrate a set of points along each camera ray.

The simple formulation of NeRF and its unprecedented rendering quality are arguably the main reasons of its popularity. However, it also suffers from long training times, since each pixel of each input view must be seen multiple times until convergence. To tackle this issue, a possible solution is to reduce the number of input views and learn a 3D scene representation from few sparse cameras. Existing few-shot methods focus on regularizing the volumetric density learned by NeRF with new loss functions \cite{infonerf,freenerf} and additional inputs \cite{dietnerf,dsnerf,densedepth,diffusionerf,regnerf}, thus introducing complexity in the pipeline. 

Furthermore, they all assume to be given a random set of viewpoints, without control on how such cameras are selected. However, in common use cases, such as object scanning from videos acquired by a user with handheld devices, the input data consist in a dense and redundant set of frames with a known acquisition trajectory. Our insight is to better exploit such information in the input views. To this end, we propose a method, called \textit{KeyNeRF}, to identify key informative samples to focus on during training. 

Firstly, we select the best input views by finding a minimal set of cameras that ensure scene coverage. Secondly, this initial set is augmented with a greedy algorithm that promotes baseline diversity. Finally, we choose the most informative pixels for each view, in terms of their local entropy in the image. Our rays selection procedure is extremely flexible, as it operates directly at input level, and it can be implemented by only changing two lines of any existing NeRF codebase. The proposed approach outperforms state-of-the-art methods on standard benchmarks in the considered scenario, while not requiring additional inputs and complex loss functions. Our contribution to existing literature is threefold:
\begin{enumerate}
    \item We present a view selection algorithm that starts from the minimal set of cameras covering the scene and iteratively adds the next best view in a greedy way.
    \item We propose to sample pixels in a given camera plane by following a probability distribution induced by the local entropy of the image. 
    \item To the best of our knowledge, our framework is the first few-shot NeRF approach that operates at input level, without requiring additional data or regularization losses.
\end{enumerate}

\section{\uppercase{Related Work}}
\label{sec:related}

\paragraph{Few-Shot NeRF} The original formulation of NeRF \cite{nerf} requires a large set of cameras to converge, thus leading to long training times. For this reason, several methods \cite{regnerf,infonerf,mixnerf,diffusionerf,dietnerf,dsnerf} have been proposed to allow learning radiance fields from few sparse views. All these approaches introduce new loss functions to regularize the underlying representation. However, such losses might be difficult to balance and they are in contrast with one of the main advantages of NeRF, which can be trained in a self-supervised way from images with a simple MSE loss. Moreover, most of them further assume to have additional inputs, such as depth measurements \cite{dsnerf,densedepth} or other pre-trained networks \cite{dietnerf,regnerf,diffusionerf}. Specifically, DS-NeRF \cite{dsnerf} and DDP-NeRF \cite{densedepth} require sparse depth to guide sampling along each ray and optimize rendered depth. DietNeRF \cite{dietnerf} enforces high-level semantic consistency between novel view renderings with pre-trained CLIP embeddings, while RegNeRF \cite{regnerf} and DiffusioNeRF \cite{diffusionerf} maximize the likelihood of a rendered patch according to a given normalizing flow or diffusion model, respectively.

\paragraph{View Selection} Another shortcoming of existing few-shot NeRF methods is that the input dataset is sampled at random, both in terms of cameras and pixels. While this is a general assumption, it is also suboptimal in typical use cases, such as object scanning from videos, since random sampling discards geometric information about the scanning trajectory. To this end, we propose a view selection algorithm that guarantees scene coverage in an optimal way and promotes baseline diversity with a greedy procedure. View selection for 3D reconstruction has been studied in literature mainly in the context of large-scale scenarios, such as city-scale \cite{iciap} or building-scale~\cite{mauro,furukawa,ladikos} reconstruction. The usual approach is to build a visibility matrix between all the possible pairs of cameras, and to employ either graph theory \cite{furukawa,ladikos} or integer linear programming \cite{mauro,iciap} to find a relevant subset of the available views. Besides targeting a different use case, such methods have several drawbacks in the considered setting. Firstly, the visibility matrix formulation assumes to have a set of sparse keypoints as input, while our approach works only with calibrated cameras. Secondly, it requires the desired number of cameras to be specified a priori, whereas KeyNeRF schedules all the views at once. Furthermore, we improve upon existing methods by explicitly and iteratively enforcing baseline diversity.

\paragraph{Rays Sampling} The aforementioned few-shot approaches treat all pixels equally for a given input image and sample a random batch at each iteration. Other methods \cite{imap,activenerf} pioneered the use of uncertainty-based sampling of rays and estimate such uncertainty online, which leads to a computational overhead. On the other hand, we propose to compute the local entropy of the image offline and to draw pixels from such distribution, which represents by definition the most informative rays.

\section{\uppercase{Method}}
\label{sec:method}

We present a framework, based on NeRF (Section \ref{sec:back}), for novel view synthesis and 3D reconstruction from a given set of $N$ calibrated cameras. We assume to have a dense and redundant set of views, such as the frames of a video acquired by a user for object scanning. Our method, named \textit{KeyNeRF}, identifies the key information in the given set of views by greedily selecting a subset of cameras (Section \ref{sec:views}) and choosing the most informative pixels within such cameras (Section \ref{sec:rays}) with entropy-based sampling. The proposed approach improves the efficiency of NeRF, while requiring minimal code changes to its implementation. Assuming NumPy imported as \texttt{np}, both cameras and rays are drawn uniformly in NeRF:

\begin{small}
\begin{Verbatim}[commandchars=\\\{\}]
pose_idx = np.random.choice(num_poses)
rays_idxs = np.random.choice(
    num_rays, size = B, p = None
)
\end{Verbatim}
\end{small}
In KeyNeRF, we simply reduce the set of input views and change the probability distribution for sampling pixels. Differences in the code are highlighted with bold characters:
\begin{small}
\begin{Verbatim}[commandchars=\\\{\}]
pose_idx = np.random.choice(\textbf{select_cams})
rays_idxs = np.random.choice(
    num_rays, size = B, p = \textbf{entropy}
)
\end{Verbatim}
\end{small}
Our rays selection procedure is extremely flexible, as it operates directly at input level and it can be seamlessly integrated with other any NeRF approach, since it is orthogonal to improvements in loss functions or field representations \cite{nsvf,tensorf,instantngp}.

\begin{figure*}[t]
     \centering
     \begin{subfigure}[b]{0.31\textwidth}
         \centering
         \includegraphics[width=\textwidth]{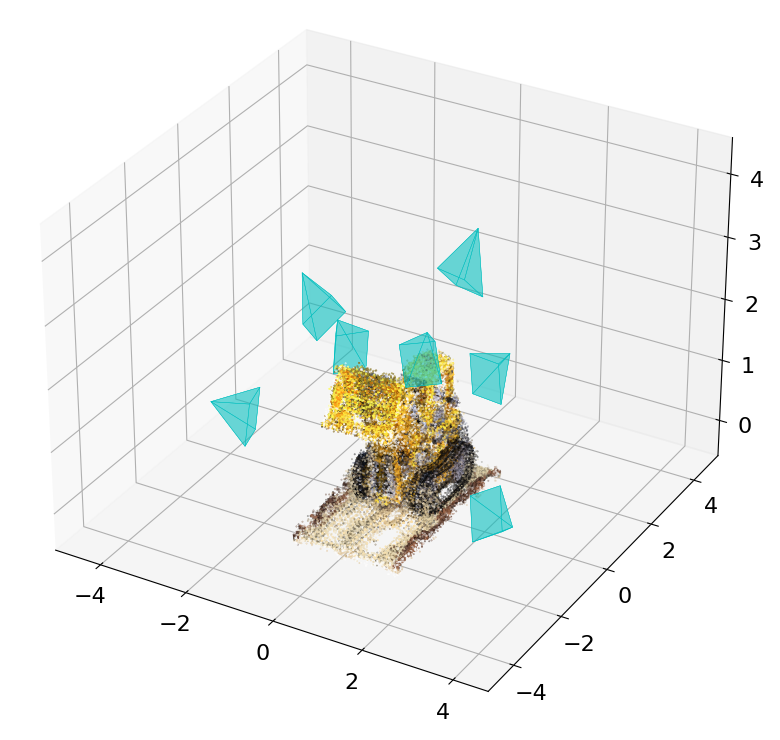}
         \label{fig:greedy}
     \end{subfigure}
     \hspace{1cm}
     \begin{subfigure}[b]{0.31\textwidth}
         \centering
         \includegraphics[width=\textwidth]{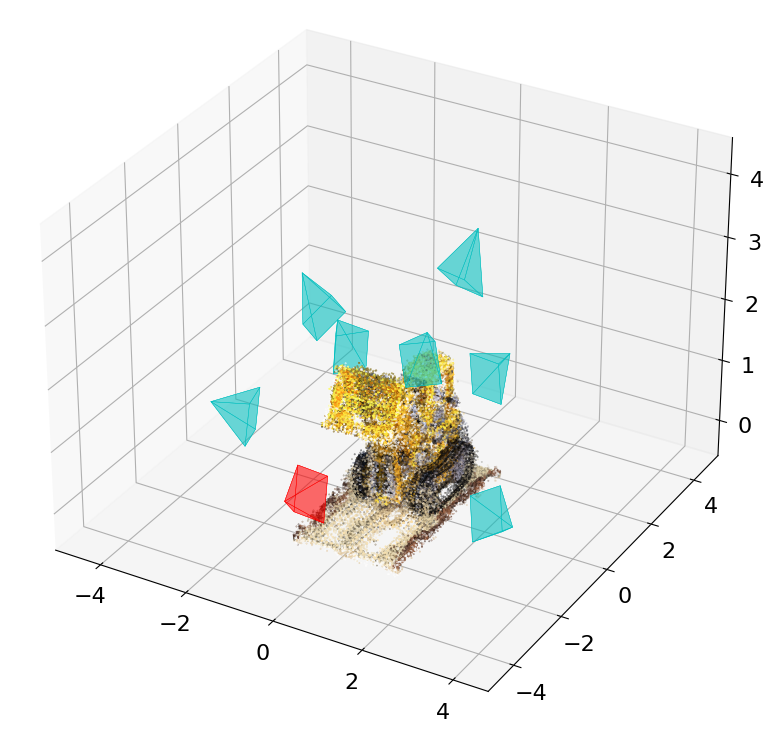}
         \label{fig:next}
     \end{subfigure}
    \caption{Illustration of the view selection procedure. The new camera (red, right) has the most diverse baseline with respect to the set of current cameras (blue, left). A proxy geometry of the scene is shown for reference.}
    \label{fig:poses}
\end{figure*}

\subsection{PRELIMINARIES}
\label{sec:back}

A Neural Radiance Field (NeRF) \cite{nerf} is a continuous and implicit representation of a 3D scene. A small MLP $F_\theta$ maps any point $\mathbf{x} \in \mathbb{R}^3$ in space and a viewing direction $\mathbf{d} \in \mathbb{S}^2$ to its corresponding density $\sigma(\mathbf{x}) \in \mathbb{R}^+$ and view-dependent color $\mathbf{c(x,d)} \in \mathbb{R}^3$. This network is trained by minimizing a reconstruction loss between the ground truth colors in the input images and the rendered colors from the radiance field. For each camera ray $\mathbf{r}(t) = \mathbf{o} + t\mathbf{d}$ with origin $\mathbf{o}$ and oriented as $\mathbf{d}$, the corresponding color $\hat{I}(\mathbf{r})$ is computed as:
\begin{equation}
\label{eq:render}
    \hat{I}(\mathbf{r}) = \int_{t_{n}}^{t_{f}} \text{exp}\left(-\int_{t_{n}}^t \sigma(s) ds \right) \cdot \sigma(t) \cdot \mathbf{c}(t) dt
\end{equation}
where $[t_{n}, t_{f}]$ is the integration boundary. In practice, both integrals are approximated by numerical quadrature with a discrete set of samples along each ray. More details can be found in \cite{nerf}. Given the ground truth color $I(\mathbf{r})$ for the ray $\mathbf{r}$, NeRF optimizes a batch of $B$ rays at each iteration. Given $N$ cameras with $M$ pixels each, an epoch needs $\frac{NM}{B}$ iterations and training requires multiple epochs. Our insight is to focus on the most informative cameras and pixels, thus significantly reducing $N$ and $M$. 

\subsection{VIEW SELECTION}
\label{sec:views}

The goal of a view selection procedure is to sample $K$ views from a dense set of $N$ available cameras $(K \ll N)$ for efficient 3D reconstruction, while (i) maintaining the visibility of the whole scene and (ii) ensuring diversity within the selected subset. Inspired by \cite{iciap}, we propose to satisfy the first constraint by solving a simple optimization problem to find the minimal set of cameras that guarantee scene coverage. Then, a greedy algorithm iteratively adds the camera with the most diverse baseline, until all the views have been scheduled.

\subsubsection{SCENE COVERAGE}

In the first phase, cameras are represented by binary variables $x_i \in \{0, 1\}$ as in \cite{iciap}, and the scene is approximated with a uniform 3D grid of $M$ points within bounds $(\mathbf{p}_{min}, \mathbf{p}_{max})$. Differently from \cite{iciap}, we do not assume to have sparse keypoints as input, but they can be added to the scene, if available. Let $\mathbf{A}_j \in \mathbb{R}^N$ be a visibility vector with elements $a_{ij} = 1$ if point $j$ is visible in camera $i$, 0 otherwise. The following integer linear programming (ILP) problem is then formulated:
\begin{equation}
\begin{array}{ll@{}ll}
\min  & \displaystyle\sum\limits_{i=1}^{N} x_{i} \\ [\bigskipamount]
\mathrm{s.t.} 
& \mathbf{A}_j^\top \mathbf{x} > 0 & \quad \forall j=1 ,\dots, M
\end{array}
\end{equation}
The set of cameras selected in this way ensures scene visibility, but some regions of interest might not be fully covered with sufficient baseline for 3D reconstruction (see Figure \ref{fig:poses}, left). 

\begin{figure*}[t]

     \centering
     \begin{subfigure}[b]{0.25\textwidth}
         \centering
         \frame{\includegraphics[width=\textwidth]{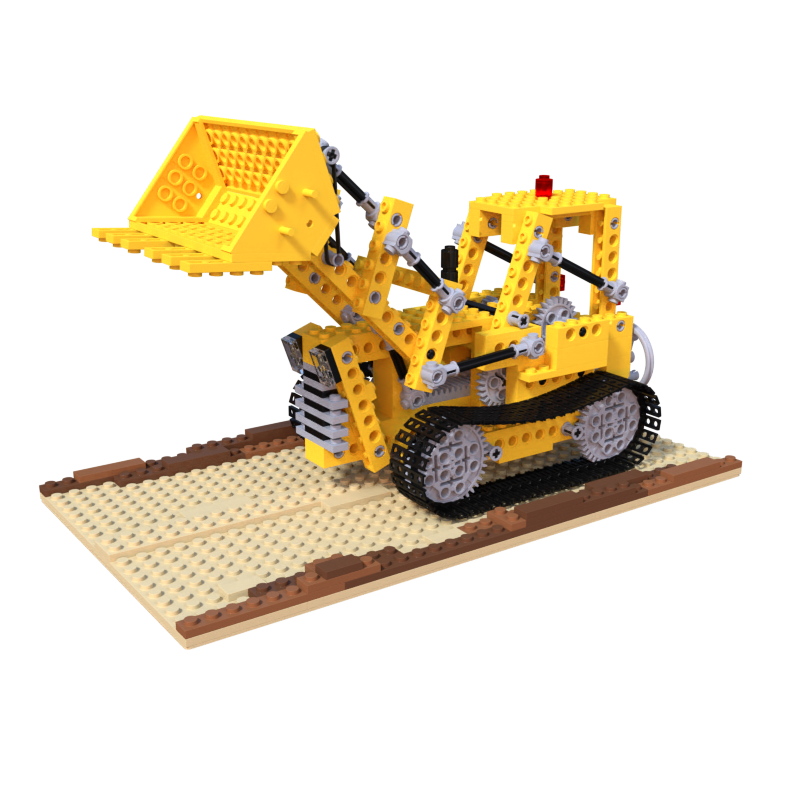}}
     \end{subfigure}
     \hspace{1.2cm}
     \begin{subfigure}[b]{0.25\textwidth}
         \centering
         \frame{\includegraphics[width=\textwidth]{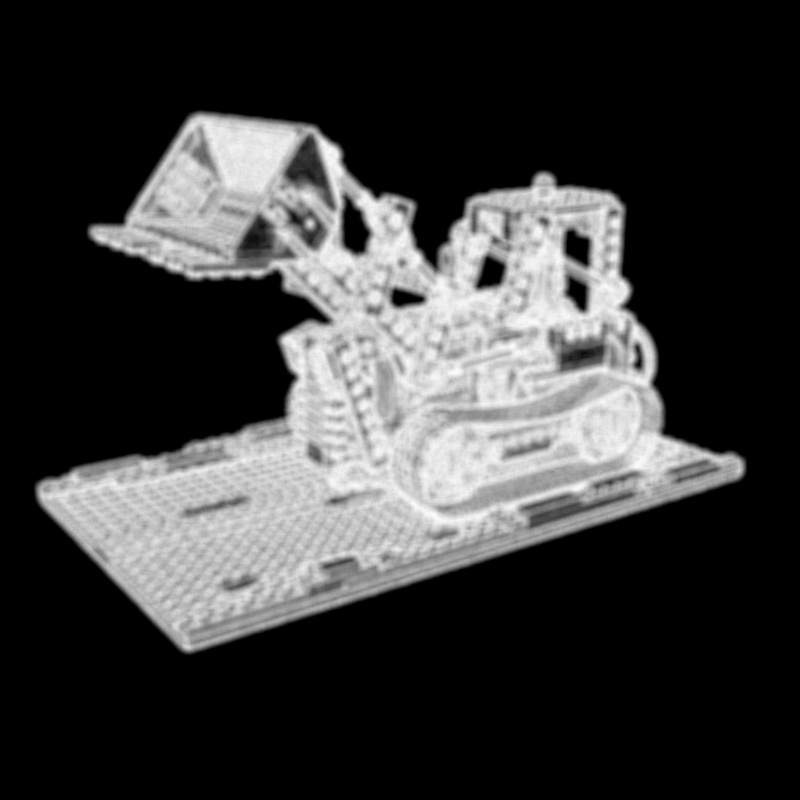}}
     \end{subfigure}
     \vspace{0.3cm}
        \caption{Probability distribution over pixels (right) for an example input image (left), induced by the local entropy of the image when sampling a batch of rays for training NeRF.}
        \label{fig:entropy}
\end{figure*}

\subsubsection{BASELINE DIVERSITY}

In order to promote baseline diversity, we design a greedy view selection algorithm to choose the next best camera among the available ones with respect to the currently selected set. We generate a $N \times N$ symmetric baseline matrix $\mathbf{B}$, where $b_{ij} = b_{ji}$ is the angle between the optical axes of cameras $i$ and $j$:
\begin{equation}
    b_{ij} = \arccos\left(\frac{\mathbf{z}_i^\top \mathbf{z}_j}{|\mathbf{z}_i|\cdot|\mathbf{z}_j|}\right)
\end{equation}
Then, at each iteration step, until the desired number of cameras has been reached, we add to the selected subset the camera with the highest relative angle with respect to \textit{all} currently selected cameras, as shown in Figure \ref{fig:poses}. Practically, for each remaining view, we query from $\mathbf{B}$ its smallest score against the selected views and add to the subset the camera with the \textit{highest} smallest score. Assuming NumPy imported as \texttt{np}, let \texttt{select\_cams} be the output of the first stage:
\begin{small}
\begin{Verbatim}[commandchars=\\\{\}]
while remaining_cams:
    sub_mat = B[remain_cams][:, select_cams]
    idx = np.argmax(np.min(sub_mat, axis = 1))
    select_cams.append(remain_cams[idx])
    remain_cams.remove(remain_cams[idx])
\end{Verbatim}
\end{small}
This formulation is different from \cite{iciap}, where matchability is a hard constraint. Moreover, the iterative nature of the greedy procedure induces an implicit ranking on the set of cameras and allows the user to choose flexibly the desired $K$. This is a significant improvement with respect to \cite{iciap}, where baseline diversity is not explicitly enforced, and a different optimization problem must be solved from scratch for different values of $K$. We will show in the experiments that any $K \geq K_{min}$ leads to good results, where $K_{min}$ is the cardinality of the minimal scene coverage set. Intuitively, more views progressively improve the performances, with diminishing returns towards the end, when cameras have large overlaps with the current set and do not add relevant information.

\subsection{RAYS SAMPLING}
\label{sec:rays}

At each training iteration, NeRF \cite{nerf} samples a pose in the dataset and a batch of $B$ pixels from such camera. Typically, rays are sampled uniformly from the whole set of available pixels. However, we observe that not all rays are equally informative about the scene. For example, the background or large textureless regions in the image could be  covered with fewer samples, exploiting the implicit smoothing bias of MLPs \cite{frequencybias}. We propose to define a probability distribution over pixels and to focus on high-frequency details during training, in order to converge faster, especially in few-shot scenarios. The amount of information of a pixel $p$ can be quantified by its local entropy: 
\begin{equation}
\label{eq:entropy}
    e(p) = - \sum_{(u,v) \in \mathcal{W}} h_{uv} \log h_{uv}
\end{equation}
where $\mathcal{W}$ defines a local window around $p$ and $h$ is the normalized histogram count. In order to allow random sampling, we normalize it to a probability distribution, which is the input to \texttt{np.random.choice()}. An example is shown in Figure \ref{fig:entropy}.

\addtolength{\tabcolsep}{3pt} 
\begin{table*}[t]

\centering
\setlength\extrarowheight{0.1pt}
\begin{tabular}{ l | c | c | c | c } 
 \toprule
 Method & PSNR $\uparrow$ & LPIPS $\downarrow$ & SSIM $\uparrow$ & Avg. $\downarrow$
 \\
 \midrule
 NeRF \cite{nerf} & 24.424 & 0.132 & 0.878 & 0.055 \\
 DietNeRF \cite{dietnerf} & 24.370 & 0.127 & 0.878 & 0.054 \\
 InfoNeRF \cite{infonerf} & 24.950 & 0.117 & 0.884 & 0.050 \\
 \midrule
 KeyNeRF (w/o entropy) & \underline{25.568} & \underline{0.109} & \underline{0.895} & \underline{0.046} \\
 KeyNeRF (ours) & \textbf{25.653} & \textbf{0.106} & \textbf{0.898} & \textbf{0.045} \\
 \bottomrule
 \end{tabular}
 \vspace{0.2cm}
 \caption{Quantitative results on the Blender dataset. Best and second results are \textbf{bold} and \underline{underlined}, respectively.}
 \label{tab:keynerf_blender_quant}
\end{table*}
\addtolength{\tabcolsep}{-3pt}

\addtolength{\tabcolsep}{3pt} 
\begin{table*}[t]

\centering
\setlength\extrarowheight{0.1pt}
\begin{tabular}{ l | c | c | c | c } 
 \toprule
 Method & PSNR $\uparrow$ & LPIPS $\downarrow$ & SSIM $\uparrow$ & Avg. $\downarrow$
 \\
 \midrule
 NeRF \cite{nerf} & 20.708 & 0.491 & 0.744  &  0.128 \\
 DietNeRF \cite{dietnerf} & 19.994 & 0.511 & 0.728  & 0.138 \\
 InfoNeRF \cite{infonerf} & 20.143 & 0.576 & 0.714 & 0.143 \\
 \midrule
 KeyNeRF (w/o entropy) & \underline{21.853} & \underline{0.470} & \underline{0.759} &  \underline{0.114} \\
 KeyNeRF (ours) & \textbf{22.183} & \textbf{0.463} & \textbf{0.762} &  \textbf{0.109} \\
 \bottomrule
 \end{tabular}
 \vspace{0.2cm}
 \caption{Quantitative results on the CO3D dataset. Best and second results are \textbf{bold} and \underline{underlined}, respectively.}
 \label{tab:keynerf_co3d_quant}
\end{table*}
\addtolength{\tabcolsep}{-3pt} 

\section{EXPERIMENTS}

\subsection{IMPLEMENTATION DETAILS}

\paragraph{Dataset} We perform our experiments on two common benchmarks. The Realistic Synthetic 360${}^\circ$ dataset \cite{nerf} contains 8 scenes of different objects with diverse materials and complex illumination, rendered by Blender from 400 random viewpoints. Moreover, we randomly select a subset of 8 scenes from the CO3D dataset \cite{co3d}, which gathers a wide collection of object-centric videos from handheld devices. Due to the unavailability of pre-trained checkpoints of baseline methods on such data, a complete evaluation on this dataset would require a huge amount of computation, well beyond what we can access. To mitigate this issue, we randomly select a subset of 8 diverse scenes, both in terms of object category and individual sequence within each category. The rationale behind this choice is to establish a real-world equivalent of the synthetic dataset, with different objects, various aspect ratios and noisy camera poses from SfM \cite{sfm}.

\begin{figure*}

     \centering
     \begin{subfigure}[b]{0.41\textwidth}
         \centering
         \includegraphics[width=\textwidth]{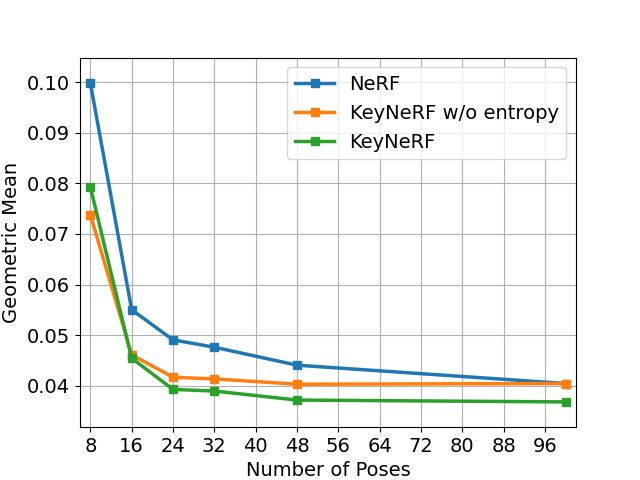}
         \caption{Ablation on the choice of $K$.}
         \label{fig:num_poses}
     \end{subfigure}
     \hspace{1cm}
     \begin{subfigure}[b]{0.41\textwidth}
         \centering
         \includegraphics[width=\textwidth]{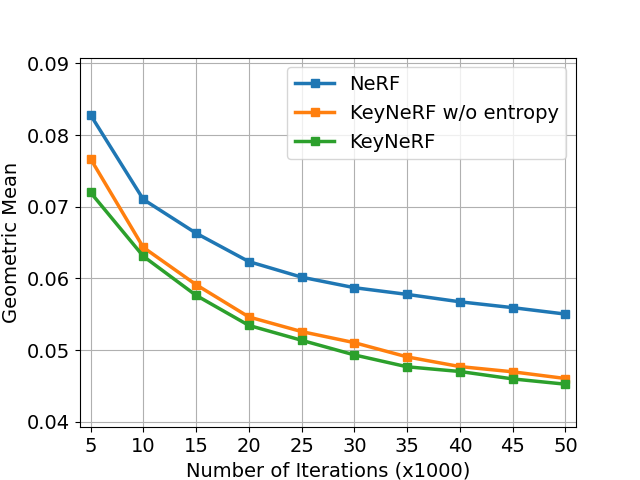}
         \caption{Ablation on the choice of $N_{iter}$.}
         \label{fig:num_iters}
     \end{subfigure}
     \vspace{0.2cm}
        \caption{Quantitative comparison between our KeyNeRF and the original NeRF \cite{nerf} as a function of the number of poses (left) and iterations (right). Lower is better.}
        \label{fig:abl}
\end{figure*}

\paragraph{Parameters} For a given batch size $B$, at each iteration, we sample $B / 2$ rays from the entropy-based distribution and $B / 2$ rays at random to ensure full coverage. All the methods are trained for $N_{iter} = 50000$ iterations with $K = 16$ poses. Note that this is a different setup than the typical few-shot scenario, where $K \leq 8$ and training is much longer ($N_{iter} \geq 200000$). We argue that this setup is overlooked in the literature, despite having significant practical relevance. In the common case of object scanning from videos, it is reasonable to assume to have more than 8 frames and the actual goal is training efficiency.  However, existing few-shot methods tend to saturate their contributions when $K > 8$ \cite{infonerf}, as shown in Table \ref{tab:keynerf_blender_quant} and Table \ref{tab:keynerf_co3d_quant}, while our method shows improved results over a wide range of values of $K$ (see Figure \ref{fig:num_poses}). The influence of both the number of poses $K$ and iterations $N_{iter}$ is ablated in Section \ref{sec:ablations}.

\paragraph{Code} The training code is based on a reference PyTorch version of NeRF \cite{nerf-pytorch}. For the rays selection procedure, we solve the ILP with the OR-Tools library \cite{ortools} and compute the image entropy with the default implementation in scikit-image.

\subsection{QUANTITATIVE RESULTS}
\label{sec:quant}

Following standard practice \cite{nerf}, we evaluate the proposed approach in terms of the image quality of novel rendered views. Such quality can be measured by three common metrics: PSNR for pixelwise differences with ground truth images, SSIM for the perceived change in structural information, and LPIPS for the similarity between the activations in a pre-trained network. Moreover, we report the geometric mean of LPIPS, $\sqrt{1 - \text{SSIM}}$ and $10^{-PSNR/10}$ to combine them in a single metric for easier comparison (reported as Avg. in Table \ref{tab:keynerf_blender_quant} and Table \ref{tab:keynerf_co3d_quant}). We compare our KeyNeRF in two different versions (i.e. with and without entropy-based rays sampling) against the original NeRF \cite{nerf} and two state-of-the-art few-shot methods \cite{dietnerf,infonerf}. Table \ref{tab:keynerf_blender_quant} and Table \ref{tab:keynerf_co3d_quant} show that both versions of KeyNeRF outperform existing approaches on synthetic and real-world data, respectively, while being much simpler to implement and more flexible to integrate with any NeRF backbone. Moreover, note that the concurrent few-shot approaches fall behind the vanilla NeRF on real-world data (see Table \ref{tab:keynerf_co3d_quant}), thus highlighting the complexity of loss weighting in such methods beyond controlled scenarios. On the other hand, KeyNeRF consistently outperforms them. 

\subsection{ABLATION STUDIES}
\label{sec:ablations}

In this section, we analyze the impact of the number of poses $K$ and the number of training iterations $N_{iter}$ on the image quality metrics, as well as the separate role of selecting views and selecting informative rays. We perform such ablations on the Blender dataset \cite{nerf}.

\begin{figure*}

     \centering
     \begin{subfigure}[b]{0.18\textwidth}
         \centering
         \includegraphics[width=\textwidth]{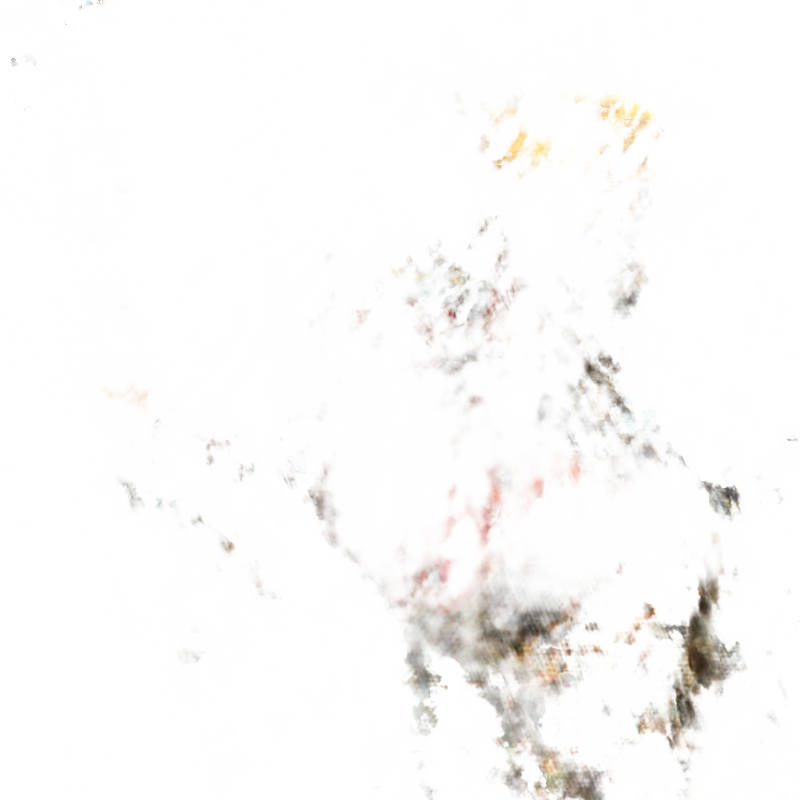}
     \end{subfigure}
     \hfill
     \begin{subfigure}[b]{0.18\textwidth}
         \centering
         \includegraphics[width=\textwidth]{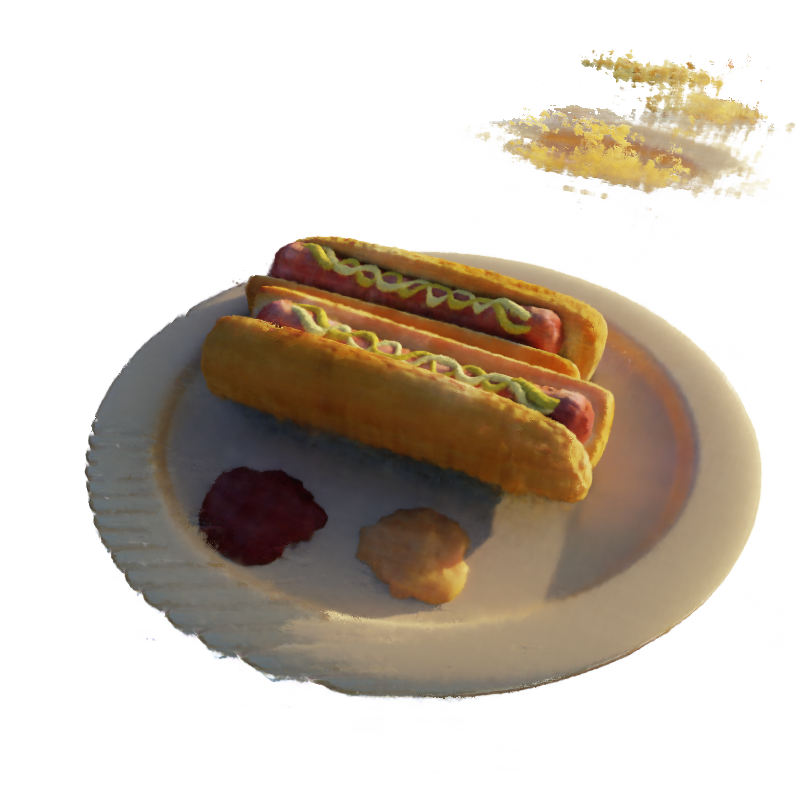}
     \end{subfigure}
     \hfill
     \begin{subfigure}[b]{0.18\textwidth}
         \centering
         \includegraphics[width=\textwidth]{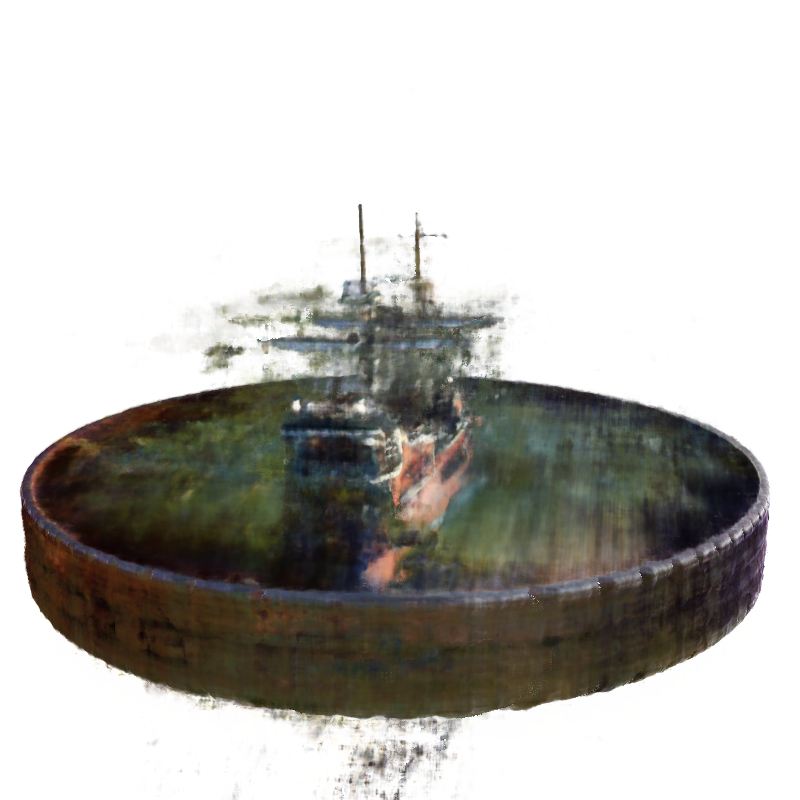}
     \end{subfigure}
     \hfill
     \begin{subfigure}[b]{0.18\textwidth}
         \centering
         \includegraphics[width=\textwidth]{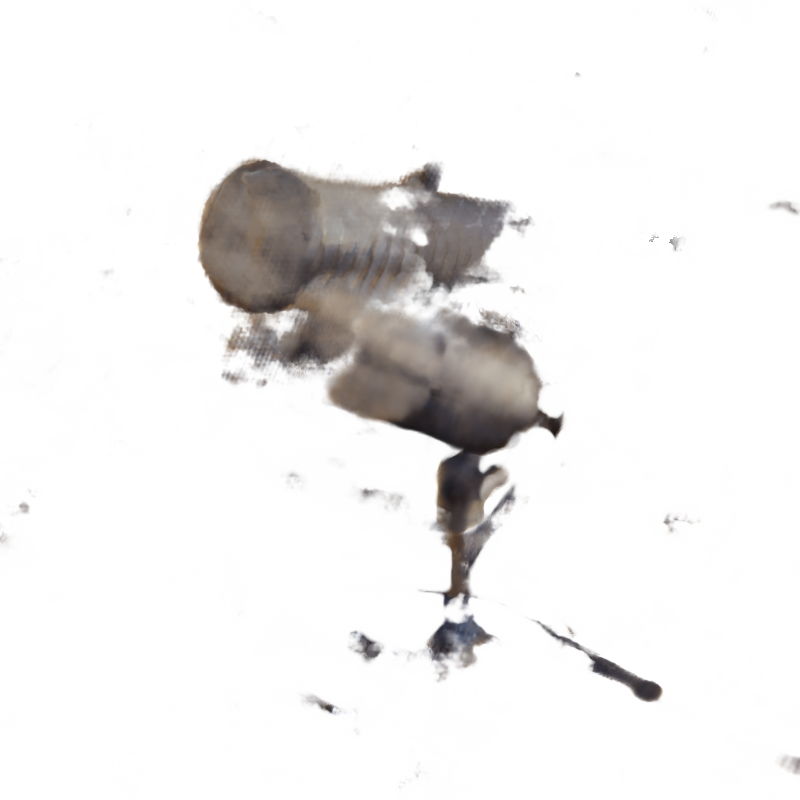}
     \end{subfigure}
     \hfill
     \begin{subfigure}[b]{0.18\textwidth}
         \centering
         \includegraphics[width=\textwidth]{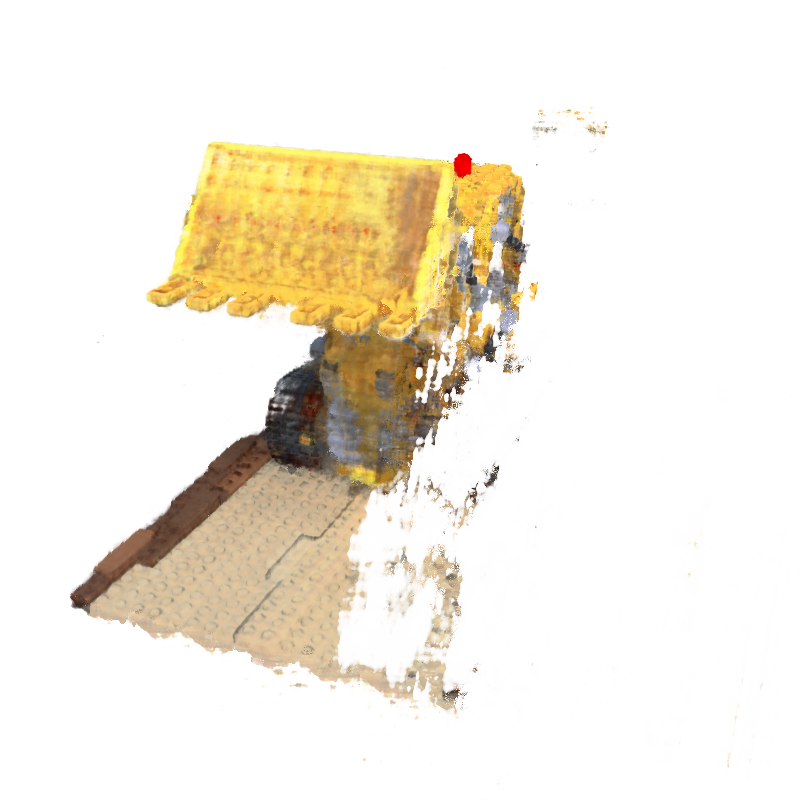}
     \end{subfigure}
     \begin{subfigure}[b]{0.18\textwidth}
         \centering
         \includegraphics[width=\textwidth]{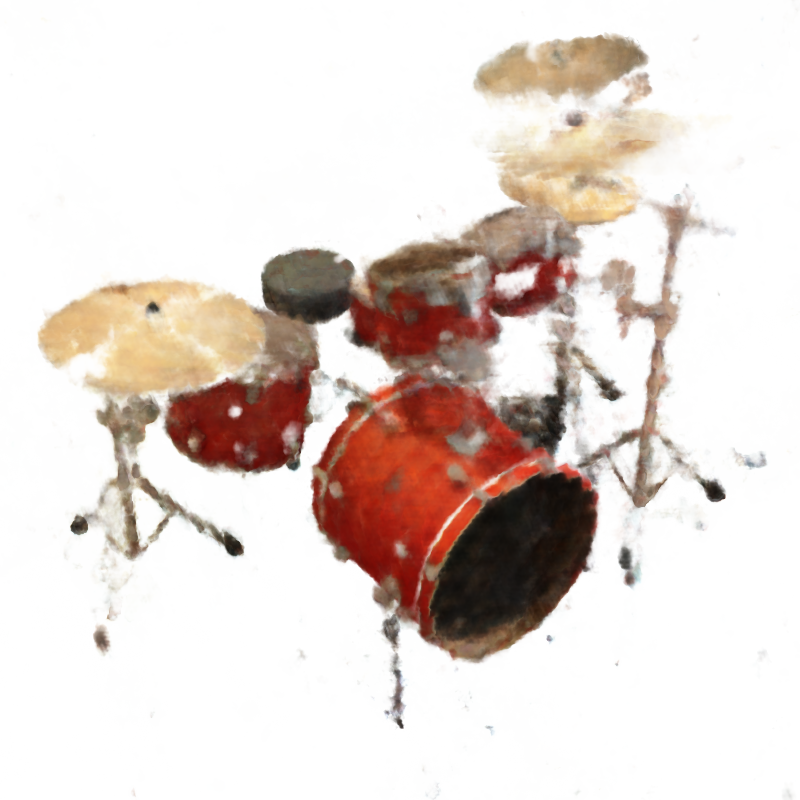}
     \end{subfigure}
     \hfill
     \begin{subfigure}[b]{0.18\textwidth}
         \centering
         \includegraphics[width=\textwidth]{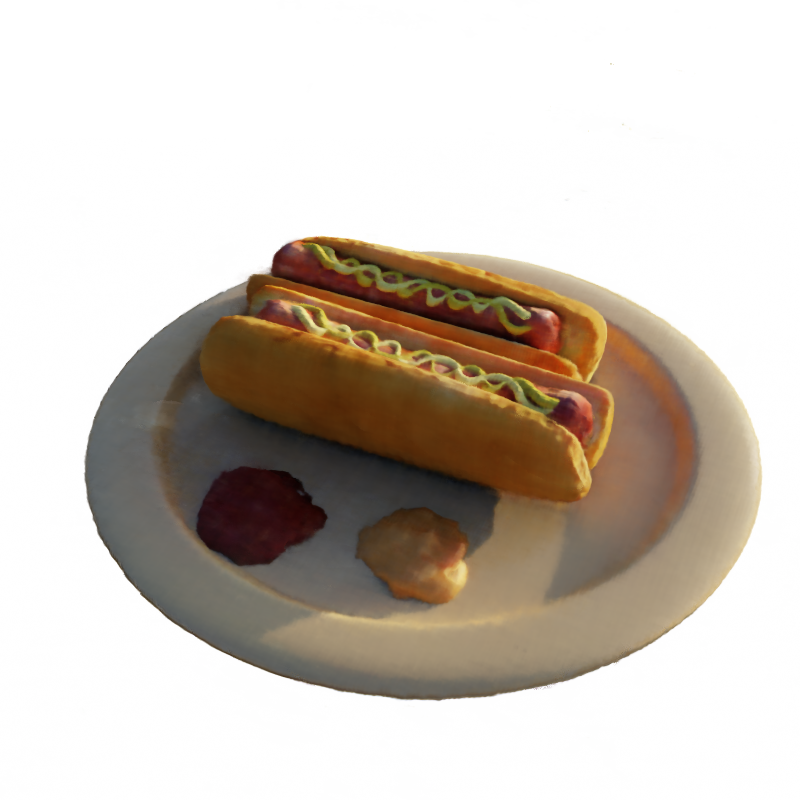}
     \end{subfigure}
     \hfill
     \begin{subfigure}[b]{0.18\textwidth}
         \centering
         \includegraphics[width=\textwidth]{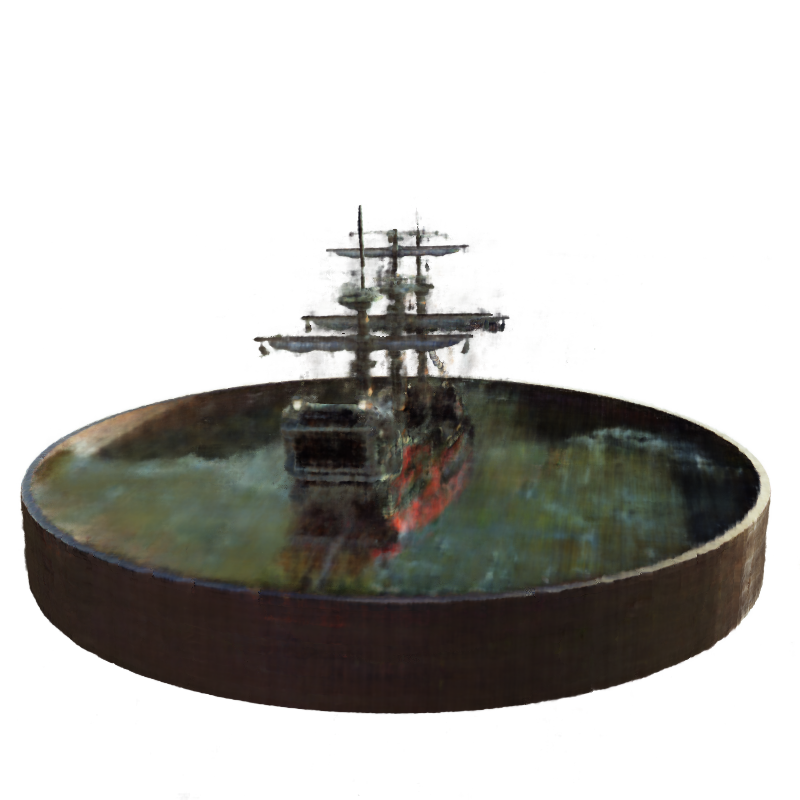}
     \end{subfigure}
     \hfill
     \begin{subfigure}[b]{0.18\textwidth}
         \centering
         \includegraphics[width=\textwidth]{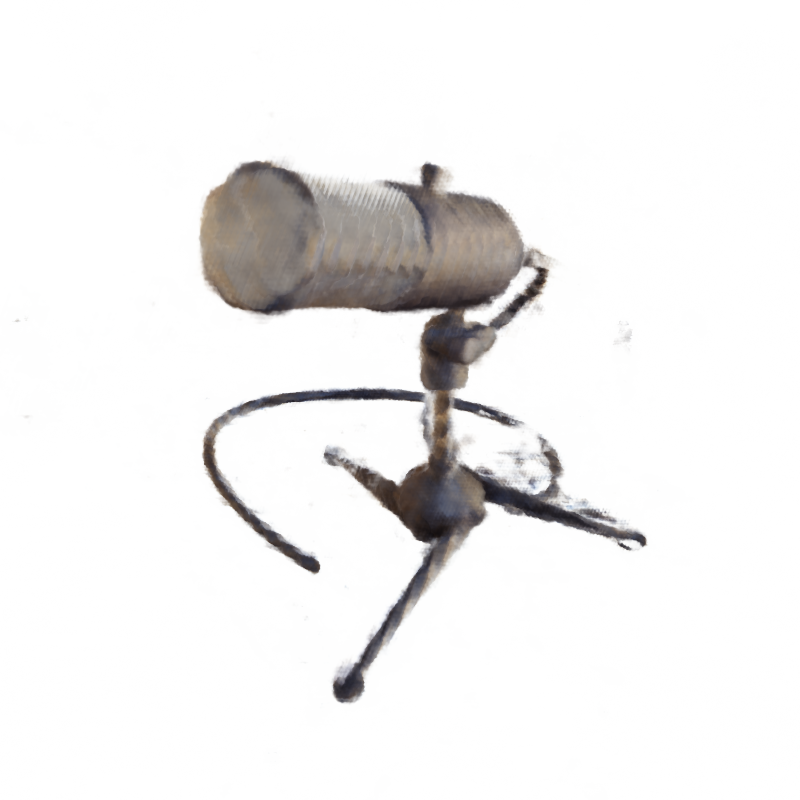}
     \end{subfigure}
     \hfill
     \begin{subfigure}[b]{0.18\textwidth}
         \centering
         \includegraphics[width=\textwidth]{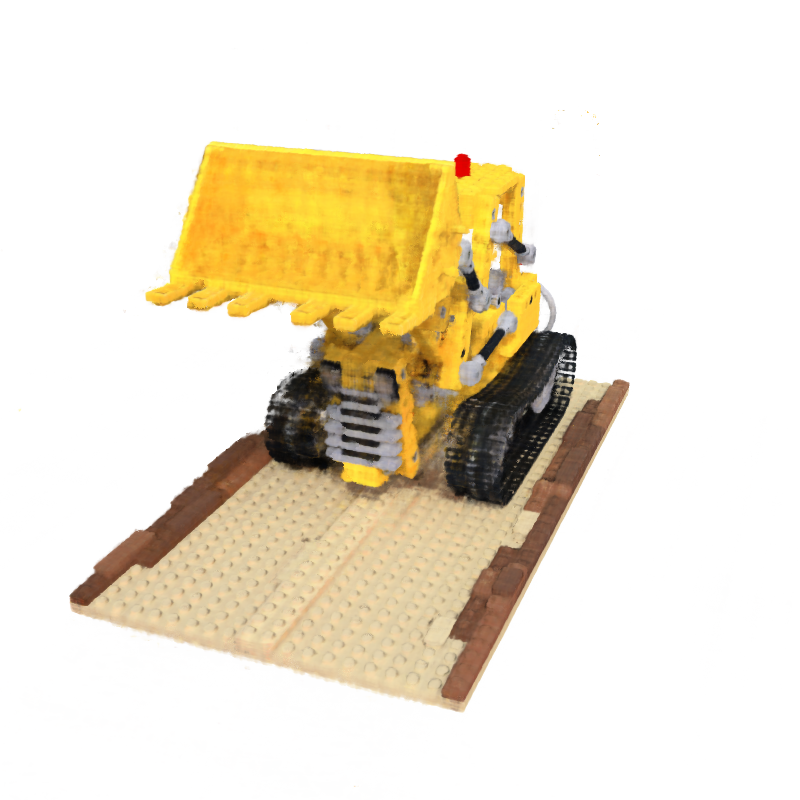}
     \end{subfigure}
    \caption{Qualitative comparison between choosing poses at random (top row) and using the proposed algorithm (bottom row) in a very few-shot setting ($K = 8$).}
    \label{fig:qual_1}
\end{figure*}

\begin{figure*}

     \centering
     \begin{subfigure}[b]{0.18\textwidth}
         \centering
         \includegraphics[width=\textwidth]{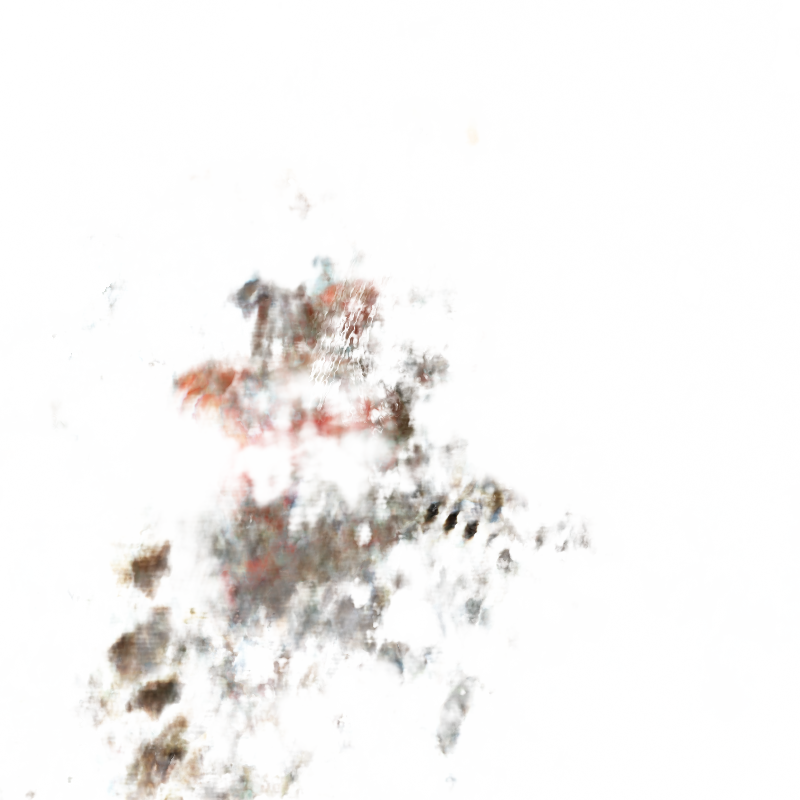}
     \end{subfigure}
     \hfill
     \begin{subfigure}[b]{0.18\textwidth}
         \centering
         \includegraphics[width=\textwidth]{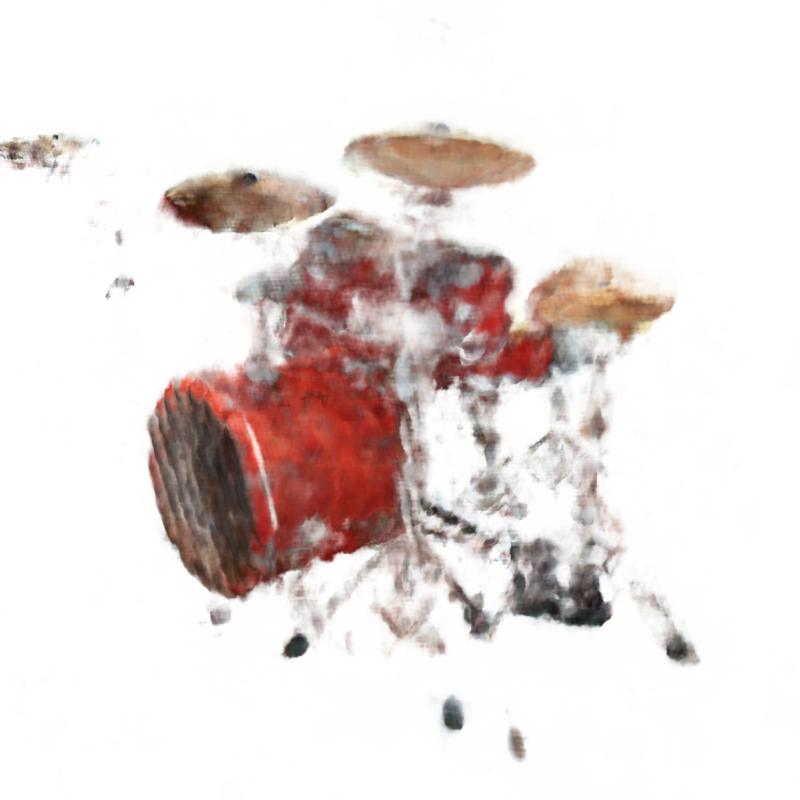}
     \end{subfigure}
     \hfill
     \begin{subfigure}[b]{0.18\textwidth}
         \centering
         \includegraphics[width=\textwidth]{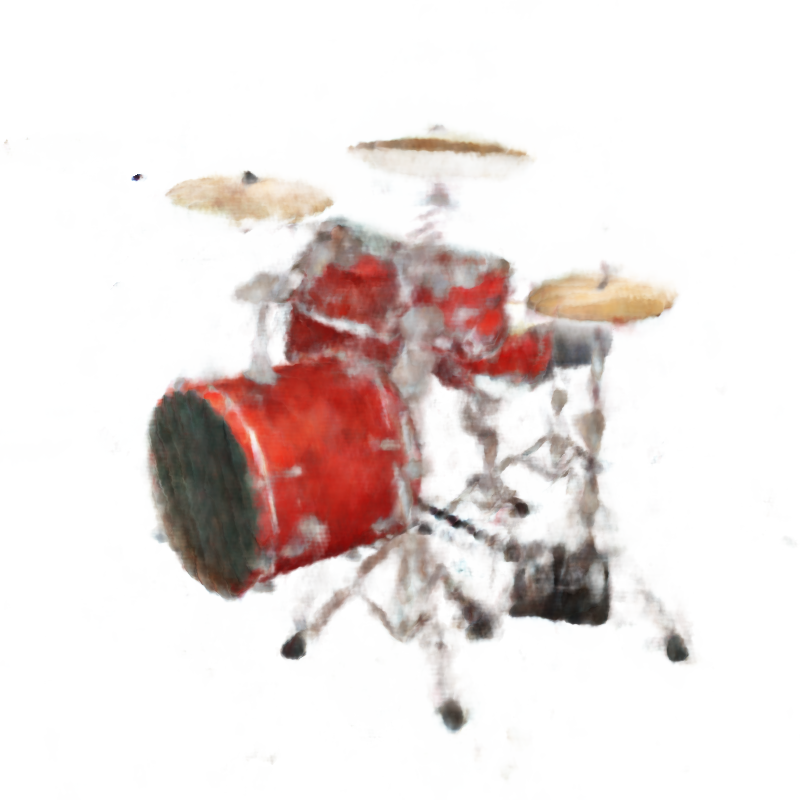}
     \end{subfigure}
     \hfill
     \begin{subfigure}[b]{0.18\textwidth}
         \centering
         \includegraphics[width=\textwidth]{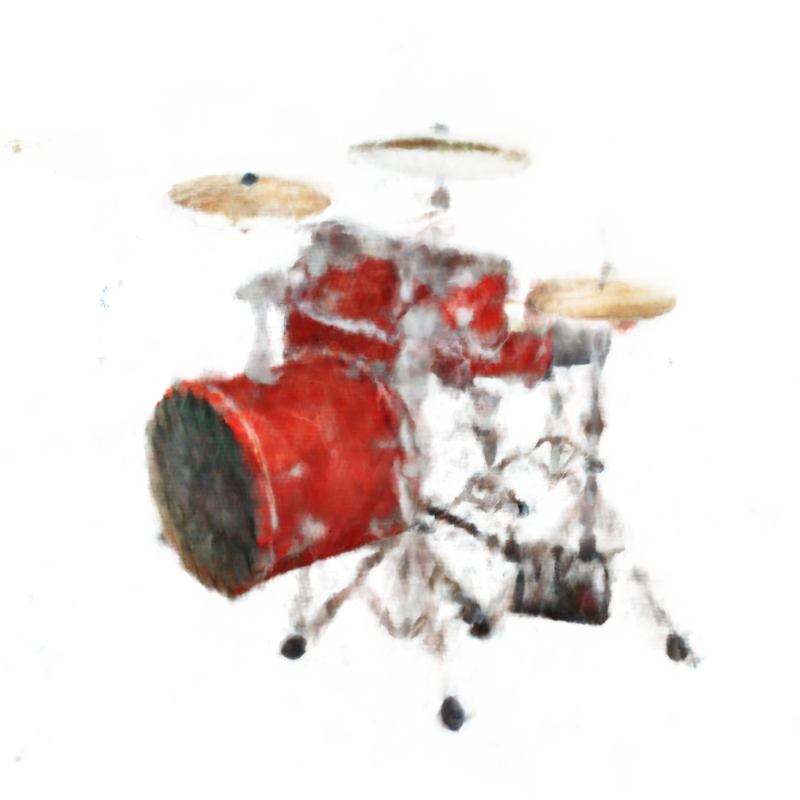}
     \end{subfigure}
     \hfill
     \begin{subfigure}[b]{0.18\textwidth}
         \centering
         \includegraphics[width=\textwidth]{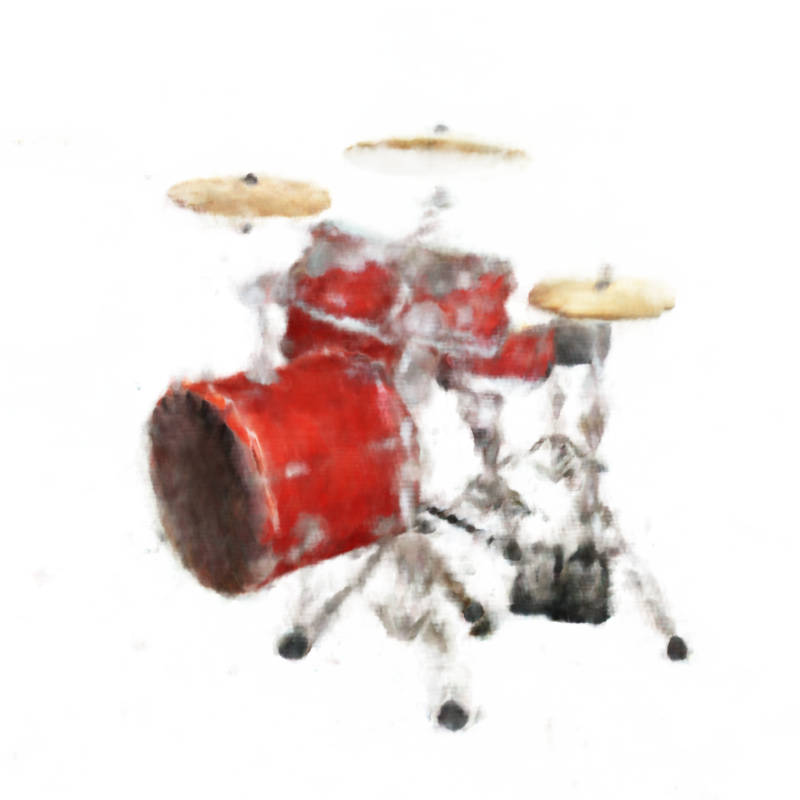}
     \end{subfigure}
     \begin{subfigure}[b]{0.18\textwidth}
         \centering
         \includegraphics[width=\textwidth]{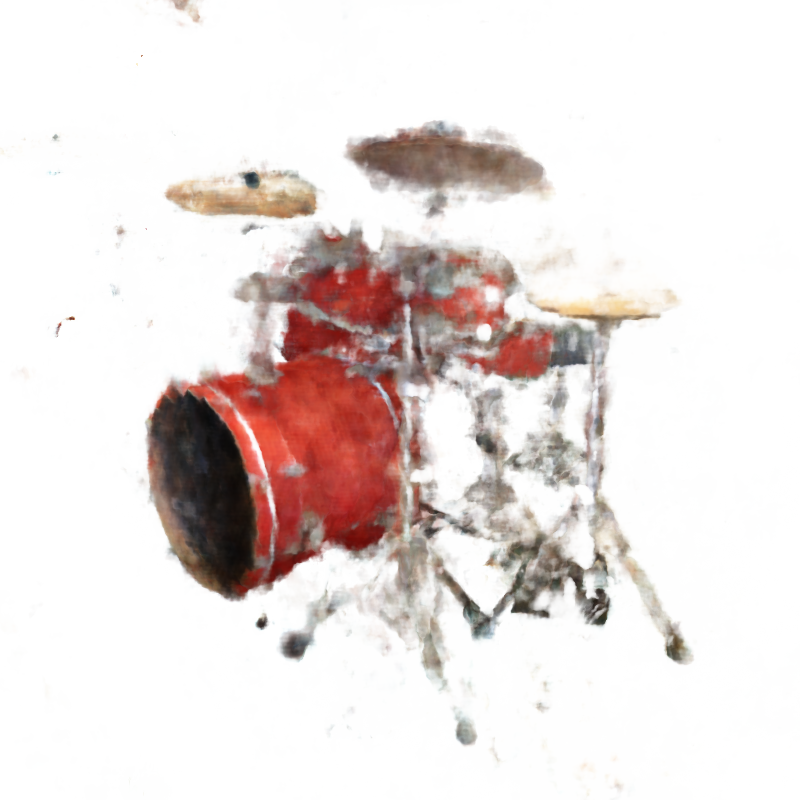}
         \caption{$K = 8$}
     \end{subfigure}
     \hfill
     \begin{subfigure}[b]{0.18\textwidth}
         \centering
         \includegraphics[width=\textwidth]{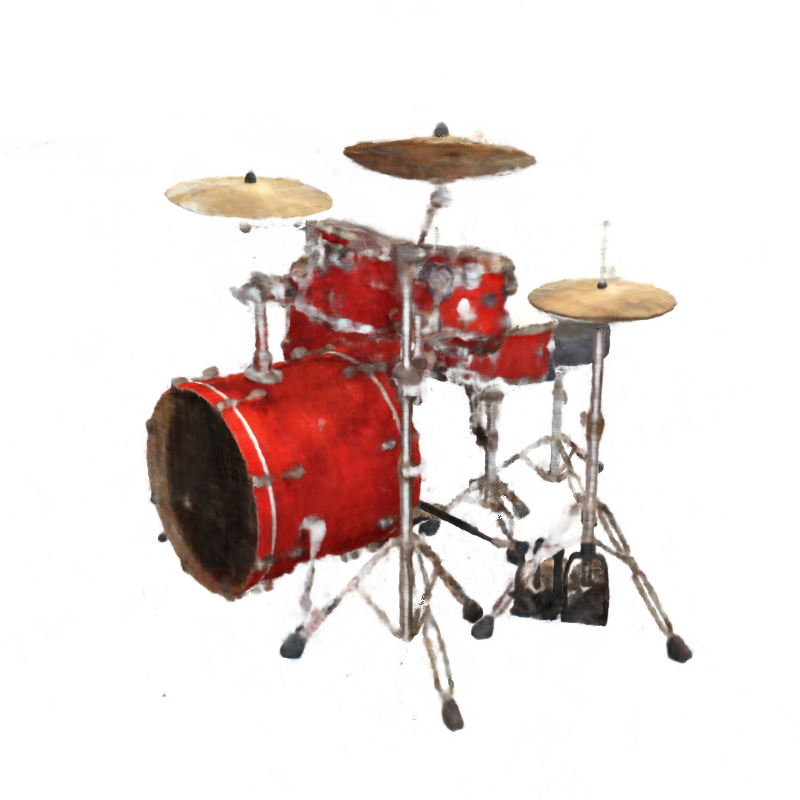}
         \caption{$K = 16$}
     \end{subfigure}
     \hfill
     \begin{subfigure}[b]{0.18\textwidth}
         \centering
         \includegraphics[width=\textwidth]{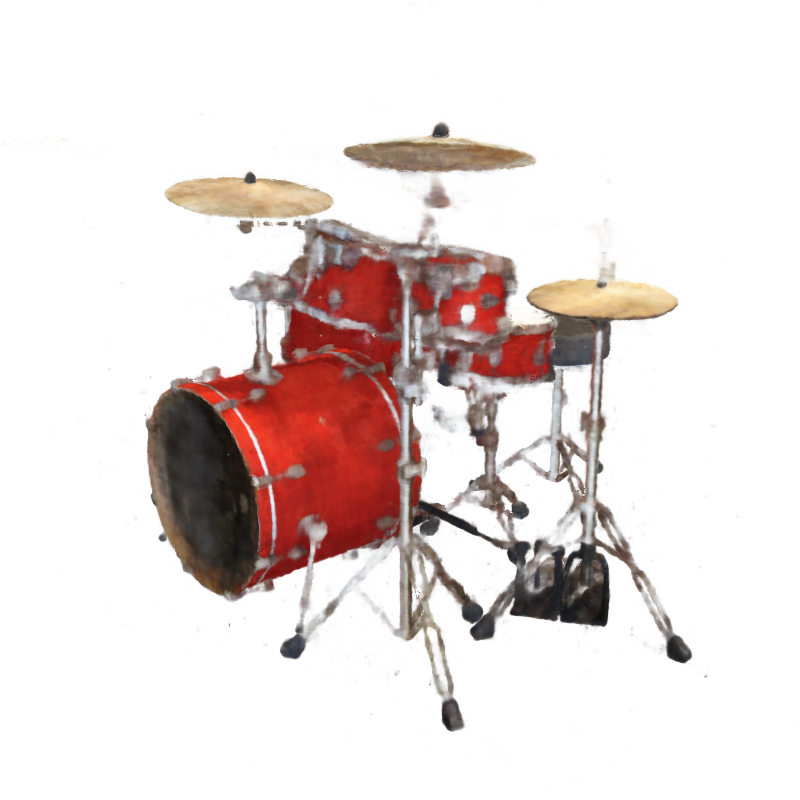}
         \caption{$K = 24$}
     \end{subfigure}
     \hfill
     \begin{subfigure}[b]{0.18\textwidth}
         \centering
         \includegraphics[width=\textwidth]{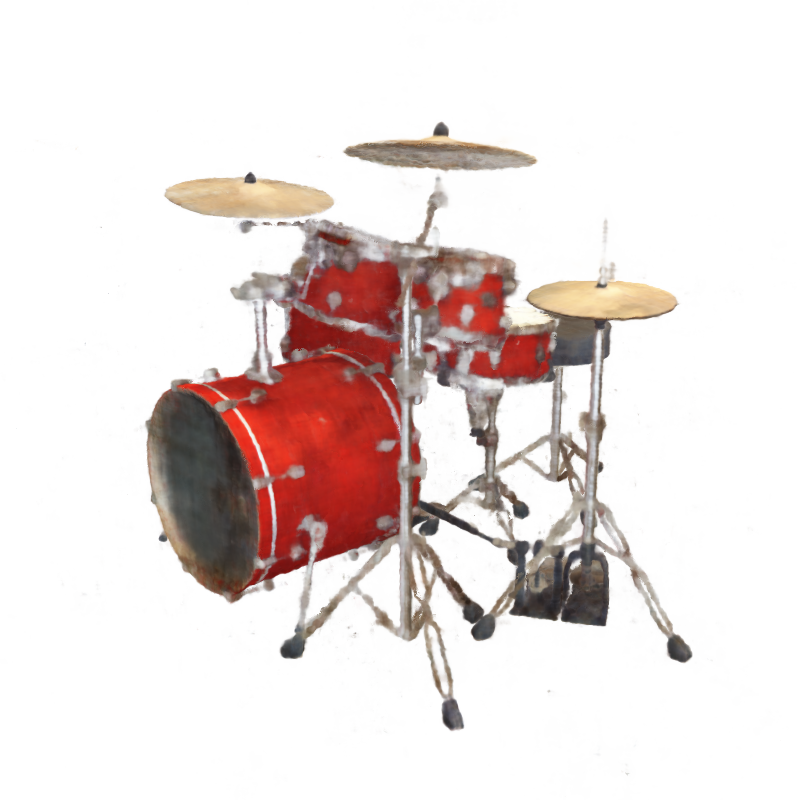}
         \caption{$K = 32$}
     \end{subfigure}
     \hfill
     \begin{subfigure}[b]{0.18\textwidth}
         \centering
         \includegraphics[width=\textwidth]{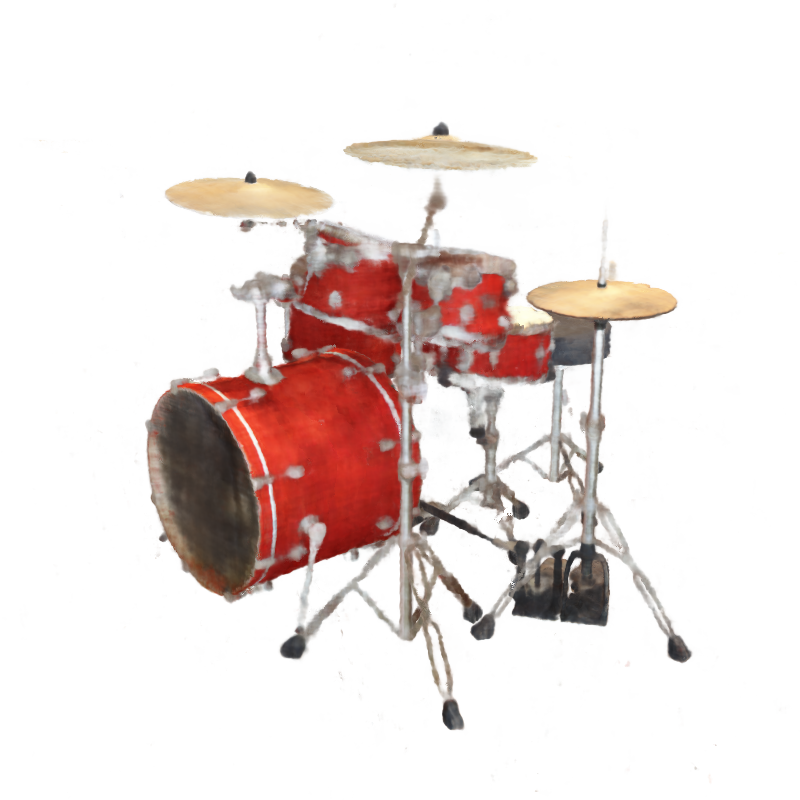}
         \caption{$K = 48$}
     \end{subfigure}
     \vspace{0.2cm}
    \caption{Qualitative comparison between choosing poses at random (top row) and using the proposed algorithm (bottom row), as a function of the number of poses $K$. Zoom in for a better view.}
    \label{fig:keynerf_qual_poses}
\end{figure*}

Figure \ref{fig:num_poses} shows that the proposed view selection method has more influence for low values of $K$ and progressively decreases, as expected. In order to clearly visualize this difference, we provide a qualitative comparison for $K = 8$ in Figure \ref{fig:qual_1} and as a function of the number of poses $K$ in Figure \ref{fig:keynerf_qual_poses}. It can be seen that our approach converges faster and with better stability. Since the coverage constraint is satisfied optimally by the view selection algorithm, KeyNeRF allows to reconstruct an approximate scene even when $K = 8$. Crucially, our improvement is still significant up to $K = 48$, whereas concurrent few-shot methods only target the lowest end of this spectrum ($K \leq 8$). 

\begin{figure*}
     \centering
     \begin{subfigure}[b]{0.18\textwidth}
         \centering
         \includegraphics[width=\textwidth]{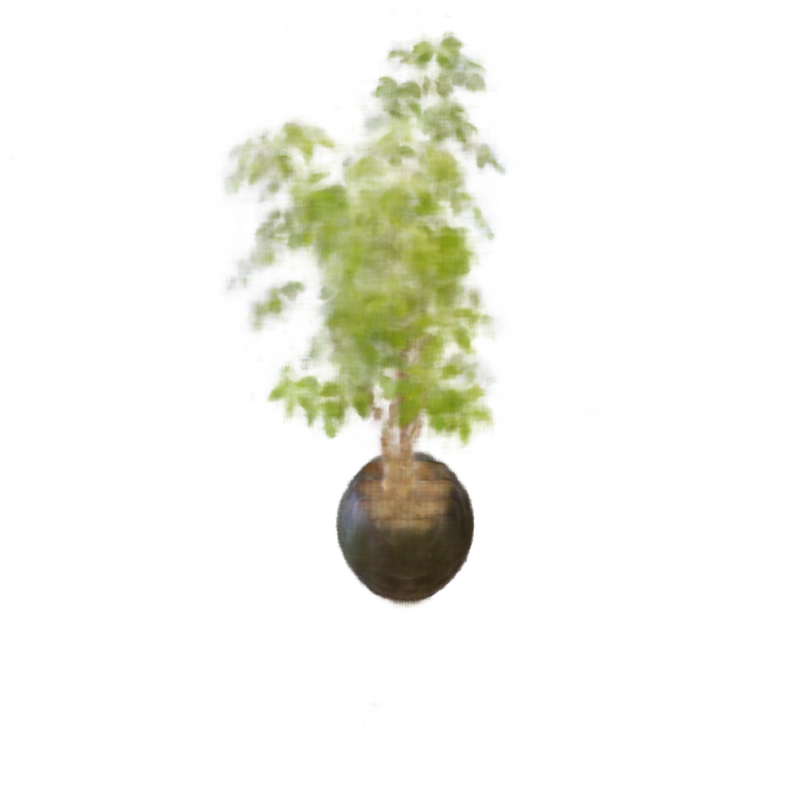}
     \end{subfigure}
     \hfill
     \begin{subfigure}[b]{0.18\textwidth}
         \centering
         \includegraphics[width=\textwidth]{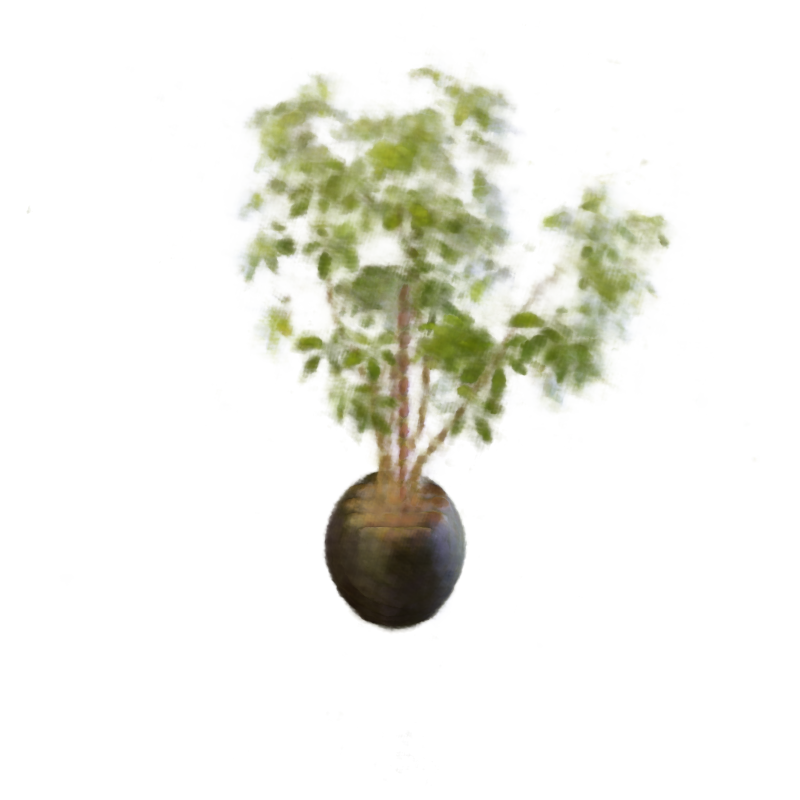}
     \end{subfigure}
     \hfill
     \begin{subfigure}[b]{0.18\textwidth}
         \centering
         \includegraphics[width=\textwidth]{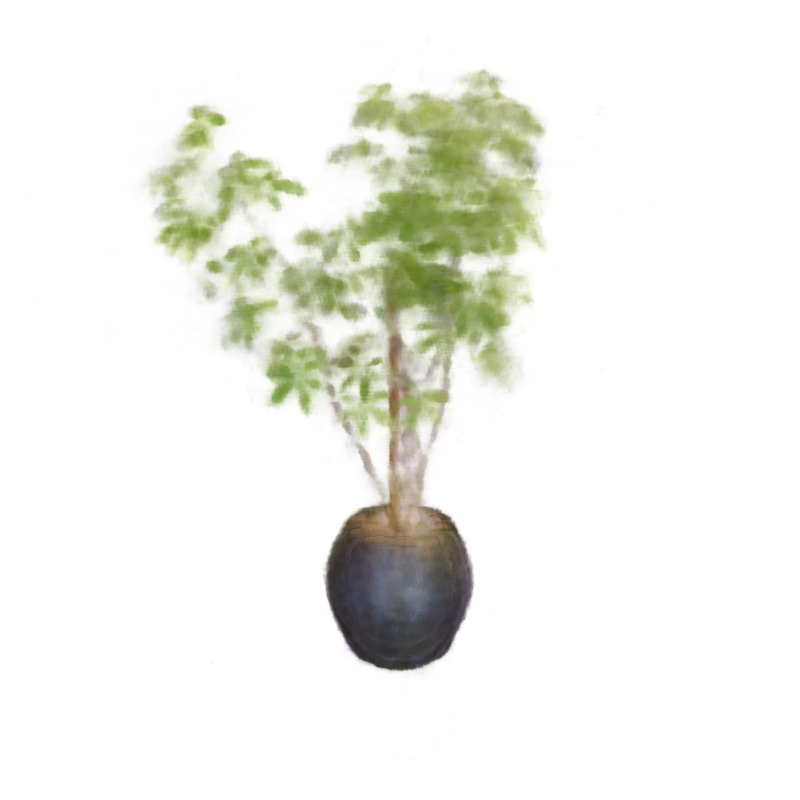}
     \end{subfigure}
     \hfill
     \begin{subfigure}[b]{0.18\textwidth}
         \centering
         \includegraphics[width=\textwidth]{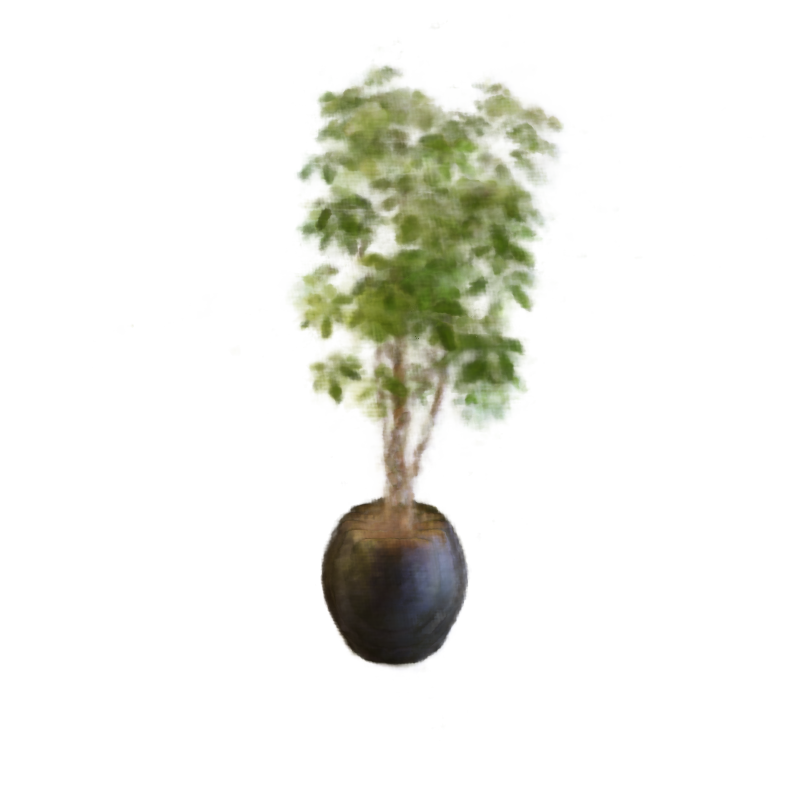}
     \end{subfigure}
     \hfill
     \begin{subfigure}[b]{0.18\textwidth}
         \centering
         \includegraphics[width=\textwidth]{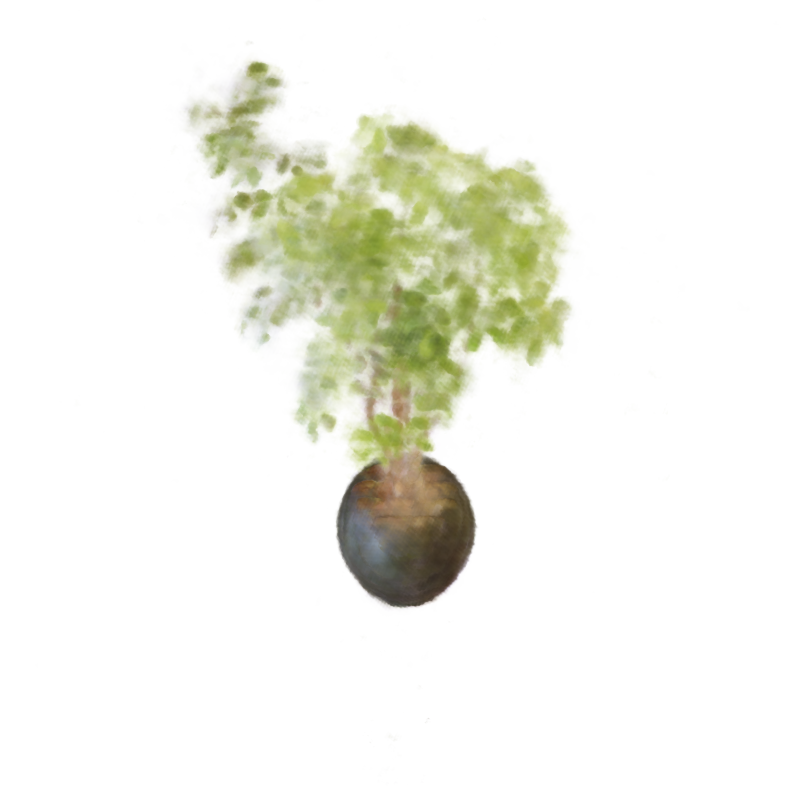}
     \end{subfigure}
     \begin{subfigure}[b]{0.18\textwidth}
         \centering
         \includegraphics[width=\textwidth]{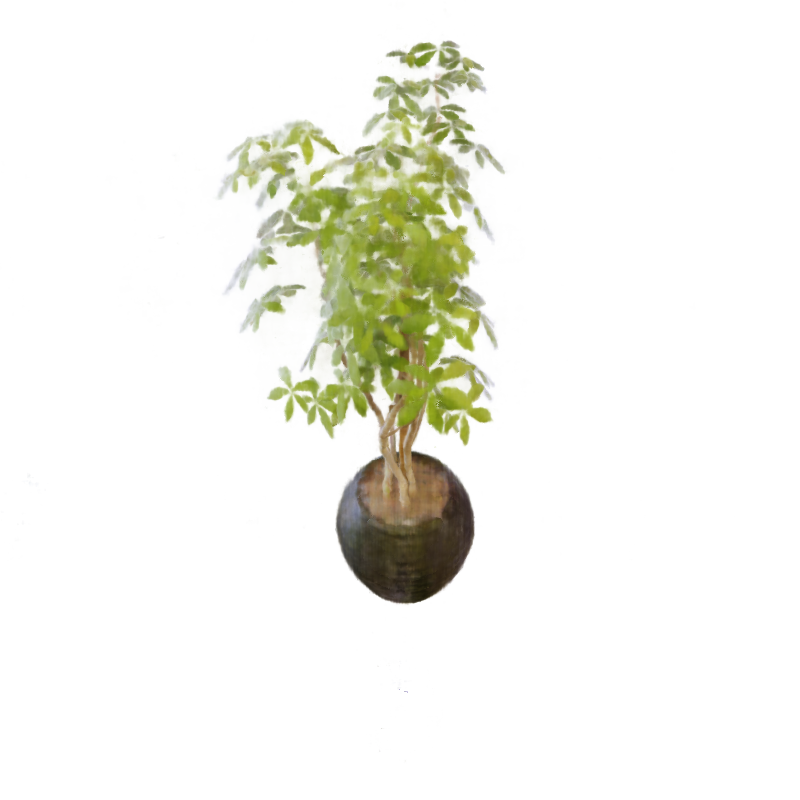}
     \end{subfigure}
     \hfill
     \begin{subfigure}[b]{0.18\textwidth}
         \centering
         \includegraphics[width=\textwidth]{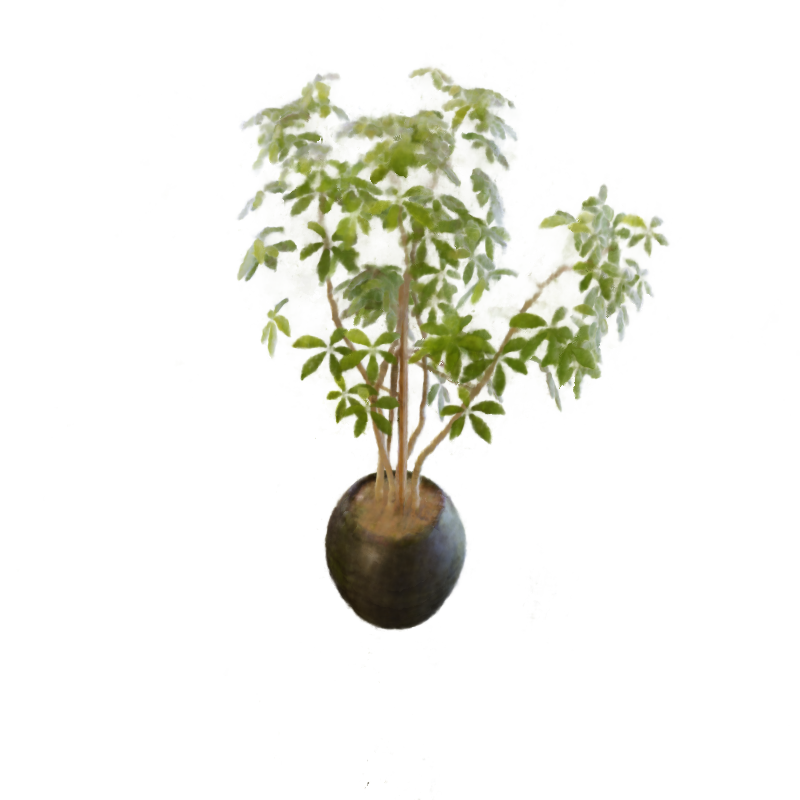}
     \end{subfigure}
     \hfill
     \begin{subfigure}[b]{0.18\textwidth}
         \centering
         \includegraphics[width=\textwidth]{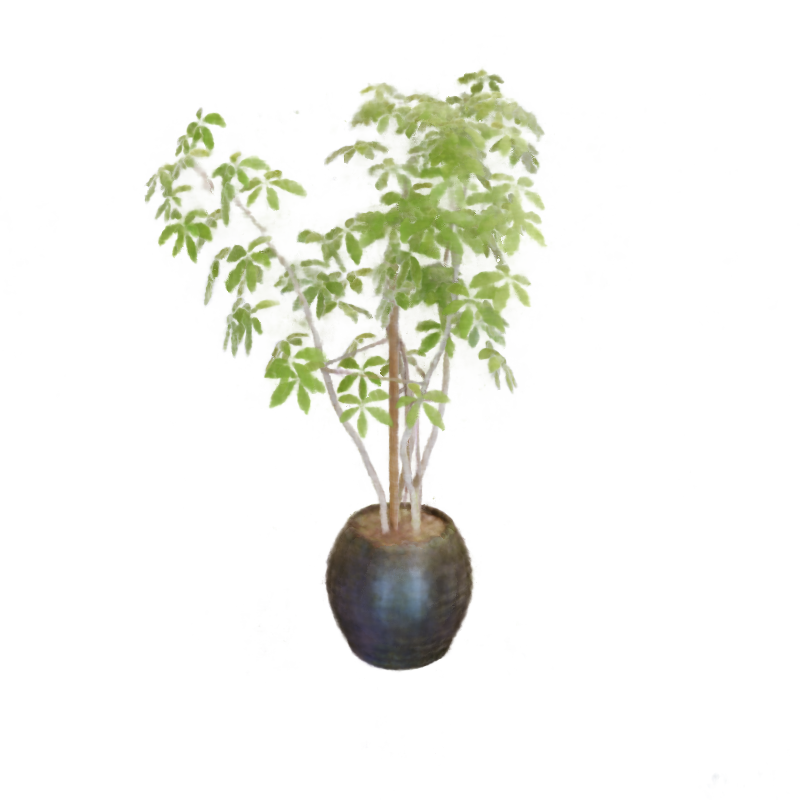}
     \end{subfigure}
     \hfill
     \begin{subfigure}[b]{0.18\textwidth}
         \centering
         \includegraphics[width=\textwidth]{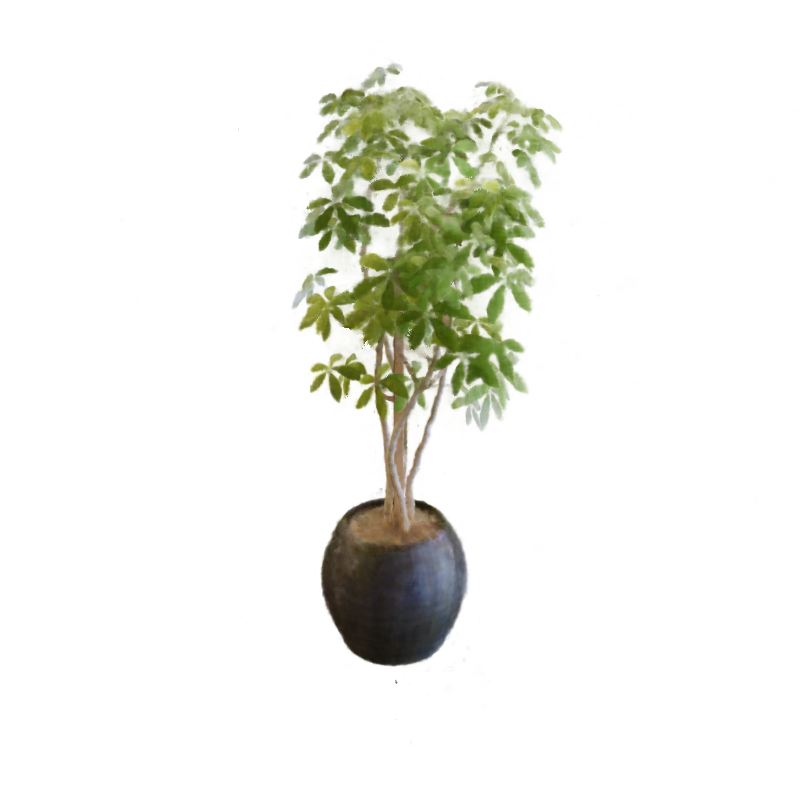}
     \end{subfigure}
     \hfill
     \begin{subfigure}[b]{0.18\textwidth}
         \centering
         \includegraphics[width=\textwidth]{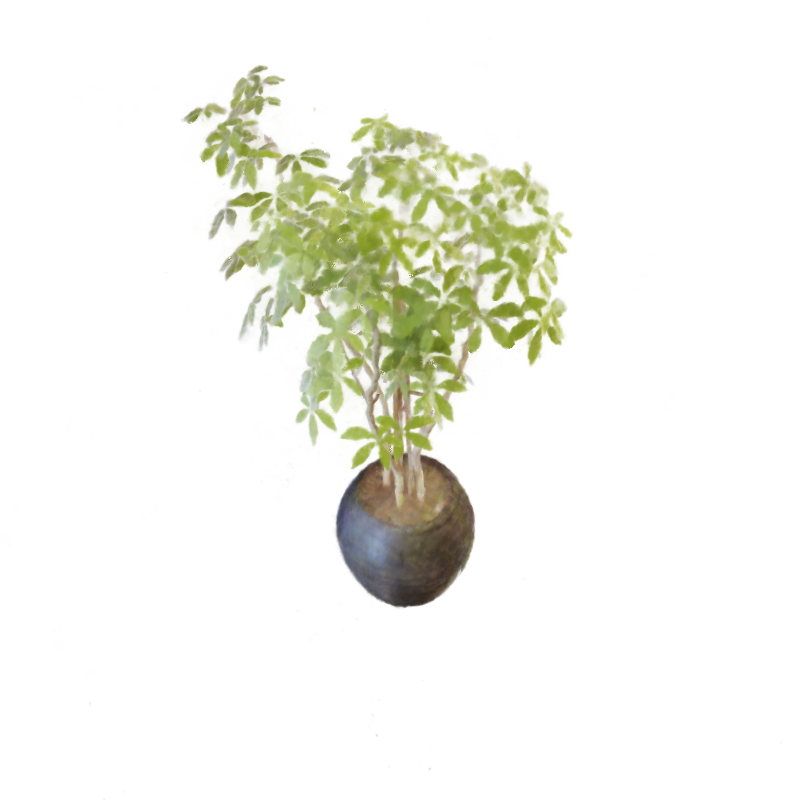}
     \end{subfigure}
    \caption{Qualitative comparison between sampling rays at random (top row) and using entropy-based sampling (bottom row) for different frames of the same scene. Zoom in for a better view.}
    \label{fig:keynerf_crisp}
\end{figure*}

Since we mainly focus on training efficiency, Figure \ref{fig:num_iters} visualizes the convergence speed in steps of 5000 iterations each. Both versions of KeyNeRF show significant improvements across the whole training runs. Moreover, note how entropy-based sampling of rays is more effective in early iterations and then saturates after around 30000 steps. This confirms that selecting the most informative rays is important, especially with a limited training budget. The quantitative results in Table \ref{tab:keynerf_blender_quant} and Table \ref{tab:keynerf_co3d_quant} underestimate the effect of this component. The lower quantitative impact is due to the fact that entropy-based sampling is most effective in fine-grained details and intricate structures, which are not well captured by numerical metrics. This is shown in Figure \ref{fig:keynerf_crisp}: sampling pixels uniformly discards crucial information, which leads to oversampling textureless areas and undersampling image regions with a lot of details. 


\begin{figure*}
     \centering
     \begin{subfigure}[b]{0.18\textwidth}
         \centering
         \includegraphics[width=\textwidth]{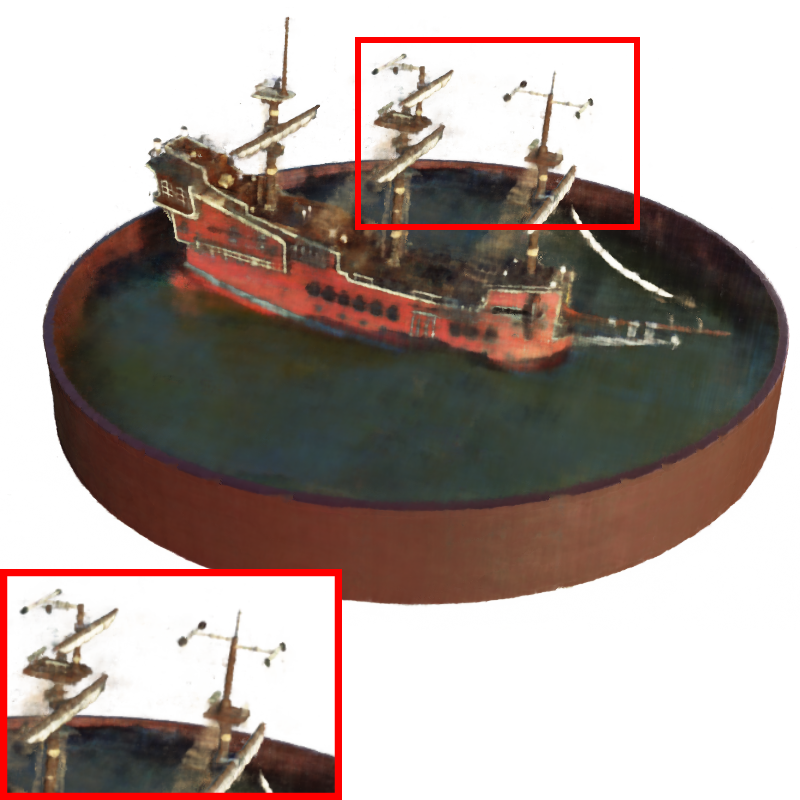}
     \end{subfigure}
     \hfill
     \begin{subfigure}[b]{0.18\textwidth}
         \centering
         \includegraphics[width=\textwidth]{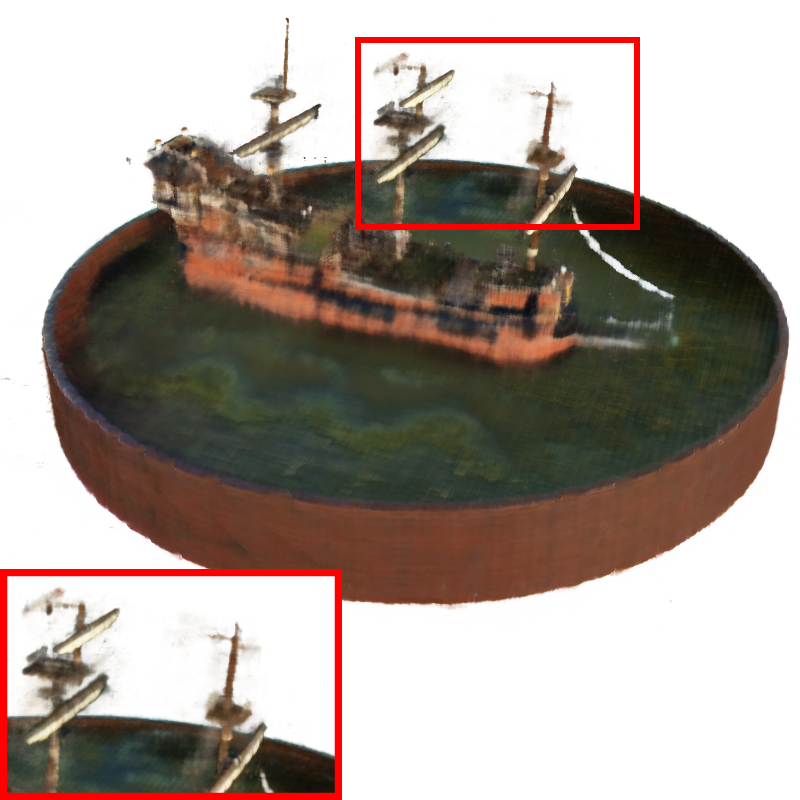}
     \end{subfigure}
     \hfill
     \begin{subfigure}[b]{0.18\textwidth}
         \centering
         \includegraphics[width=\textwidth]{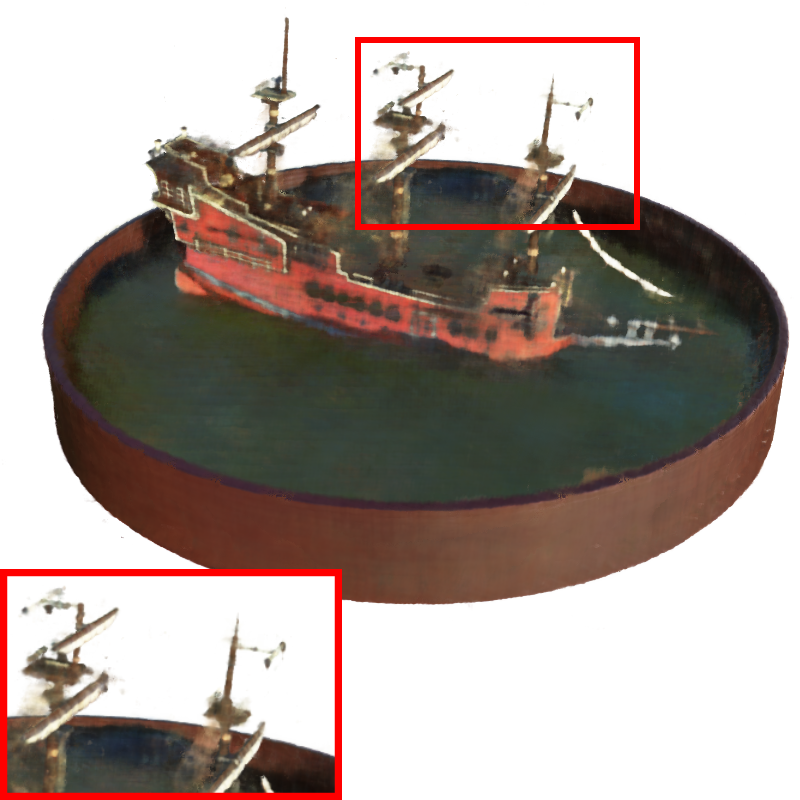}
     \end{subfigure}
     \hfill
     \begin{subfigure}[b]{0.18\textwidth}
         \centering
         \includegraphics[width=\textwidth]{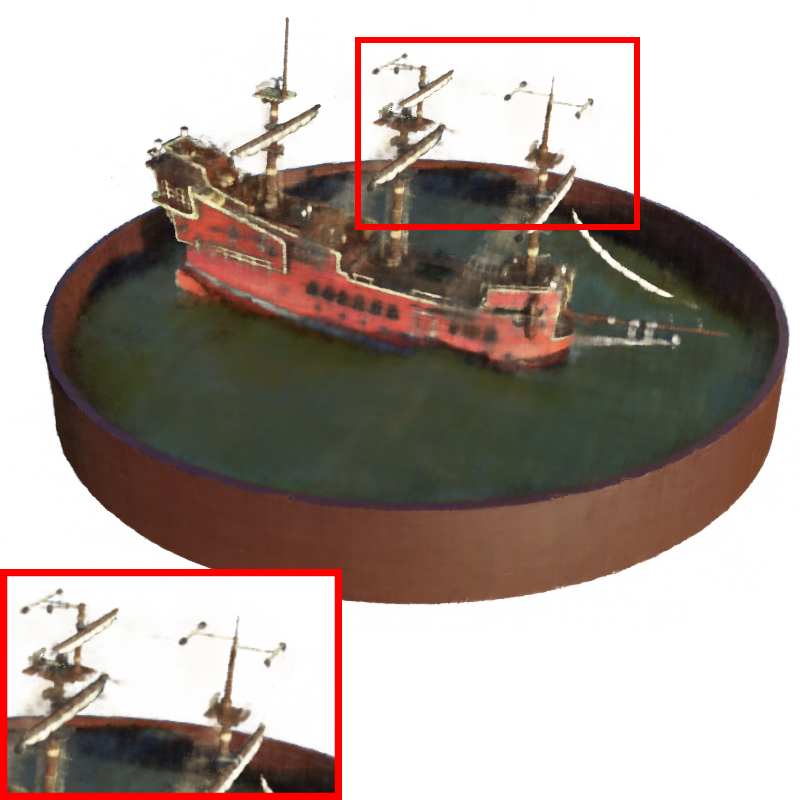}
     \end{subfigure}
     \hfill
     \begin{subfigure}[b]{0.18\textwidth}
         \centering
         \includegraphics[width=\textwidth]{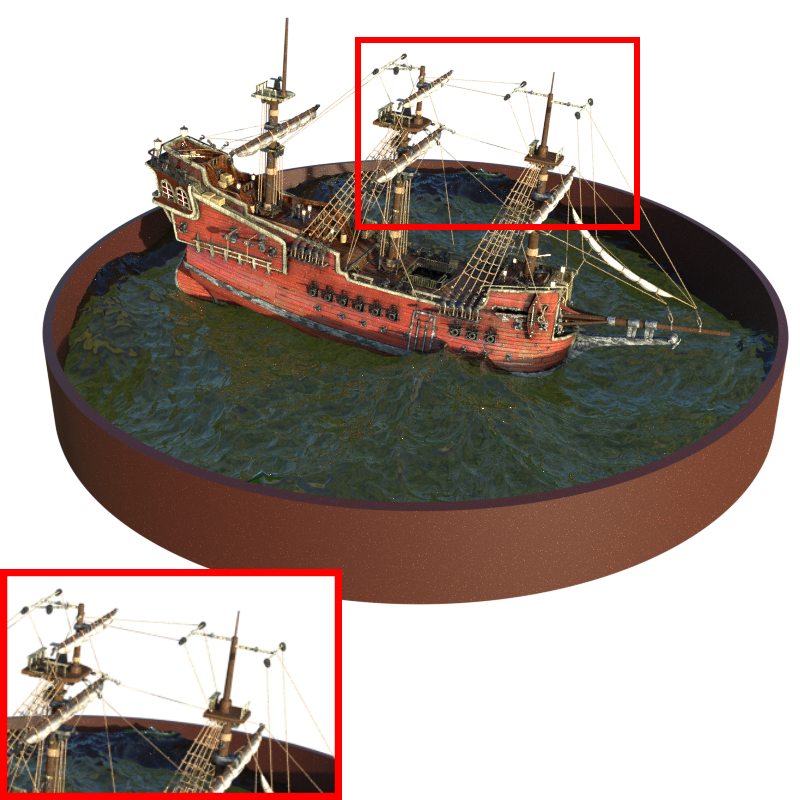}
     \end{subfigure}
     \par\smallskip
     \begin{subfigure}[b]{0.18\textwidth}
         \centering
         \includegraphics[width=\textwidth]{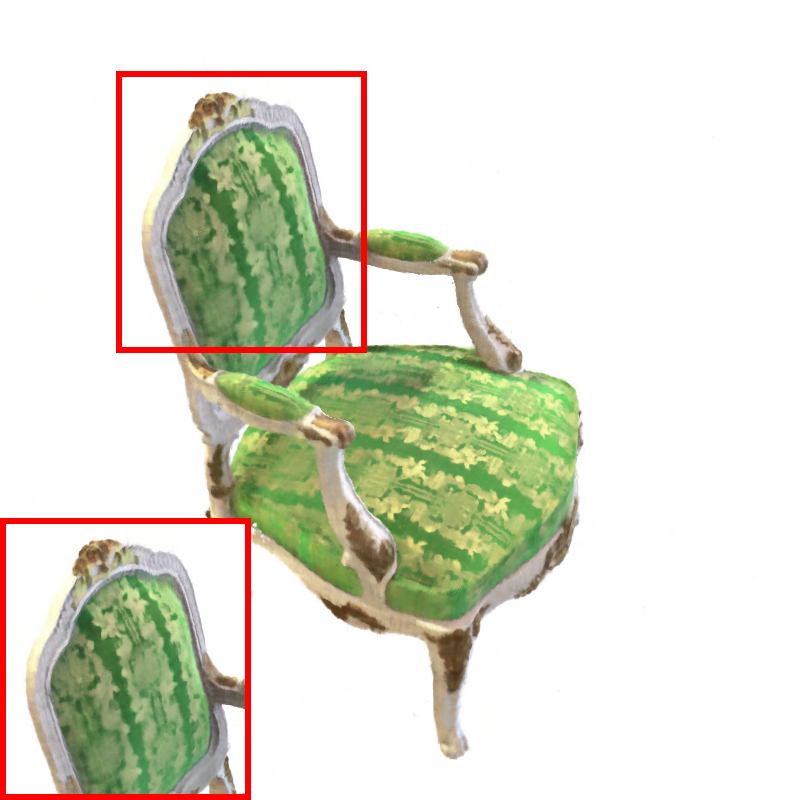}
     \end{subfigure}
     \hfill
     \begin{subfigure}[b]{0.18\textwidth}
         \centering
         \includegraphics[width=\textwidth]{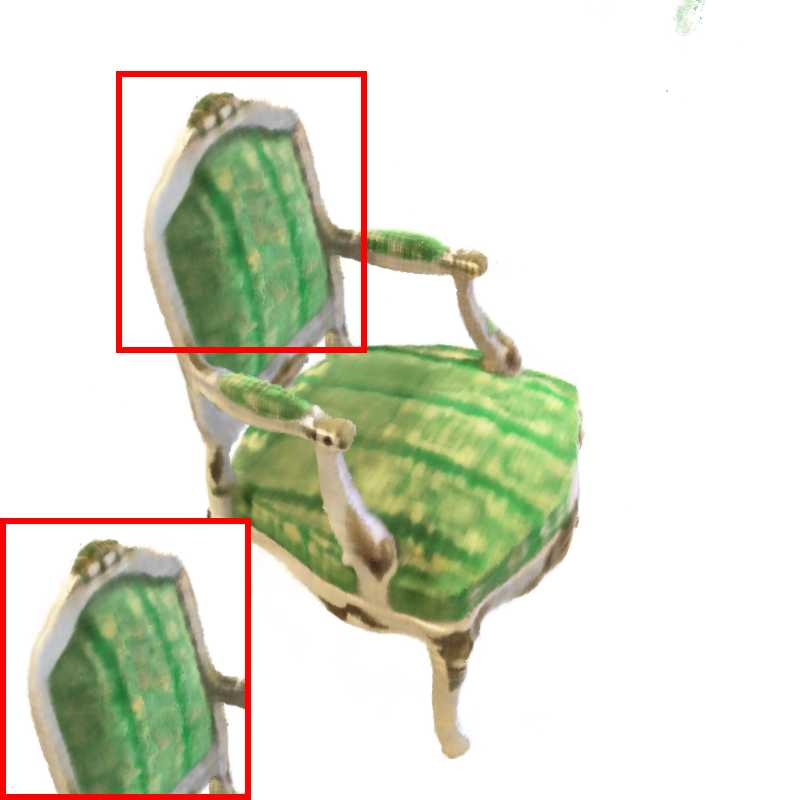}
     \end{subfigure}
     \hfill
     \begin{subfigure}[b]{0.18\textwidth}
         \centering
         \includegraphics[width=\textwidth]{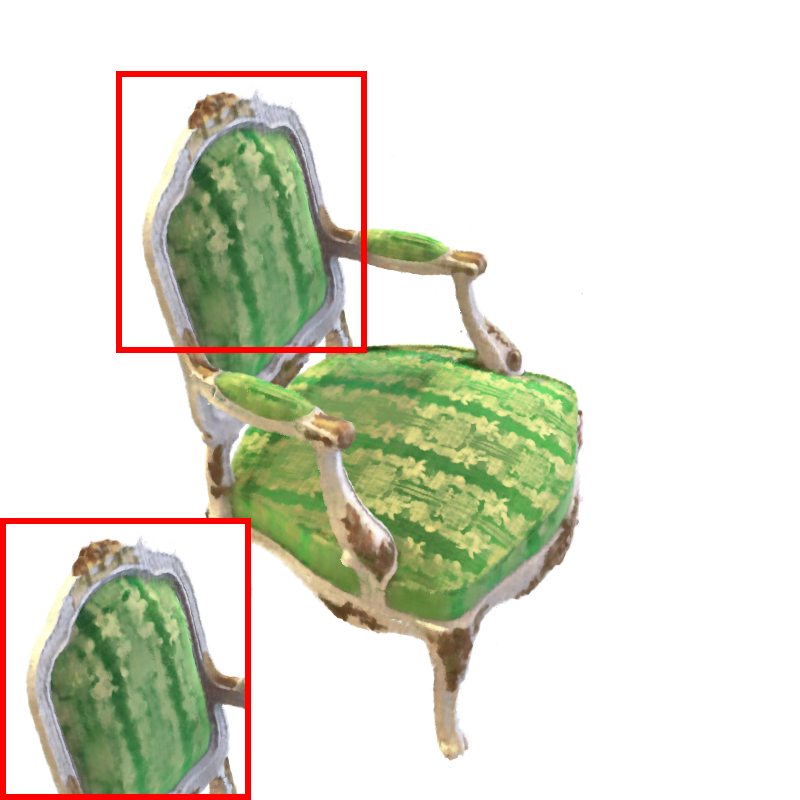}
     \end{subfigure}
     \hfill
     \begin{subfigure}[b]{0.18\textwidth}
         \centering
         \includegraphics[width=\textwidth]{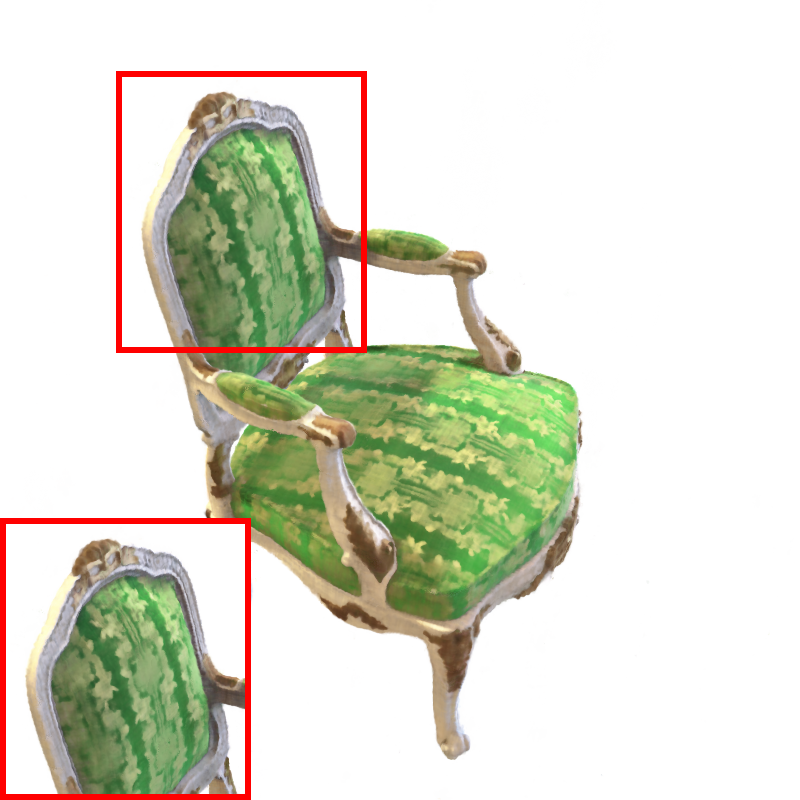}
     \end{subfigure}
     \hfill
     \begin{subfigure}[b]{0.18\textwidth}
         \centering
         \includegraphics[width=\textwidth]{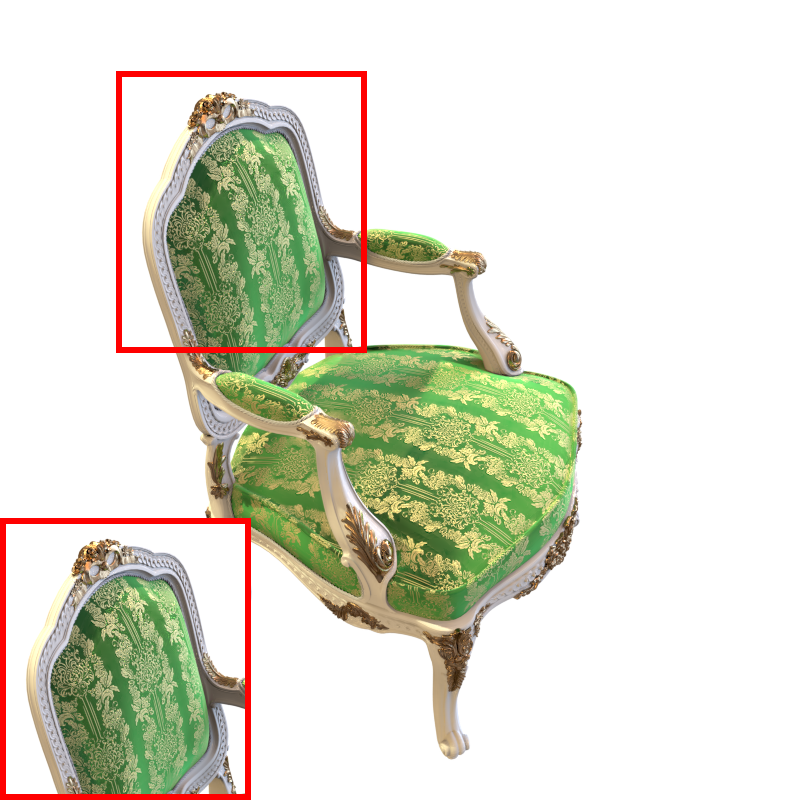}
     \end{subfigure}
     \par\smallskip
     \begin{subfigure}[b]{0.18\textwidth}
         \centering
         \includegraphics[width=\textwidth]{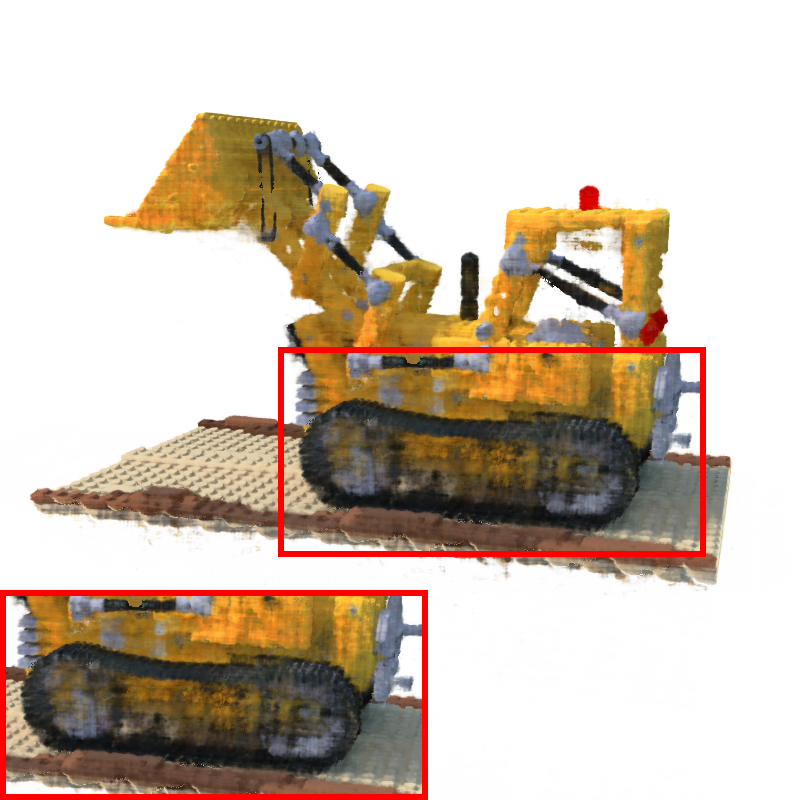}
         \caption{NeRF}
     \end{subfigure}
     \hfill
     \begin{subfigure}[b]{0.18\textwidth}
         \centering
         \includegraphics[width=\textwidth]{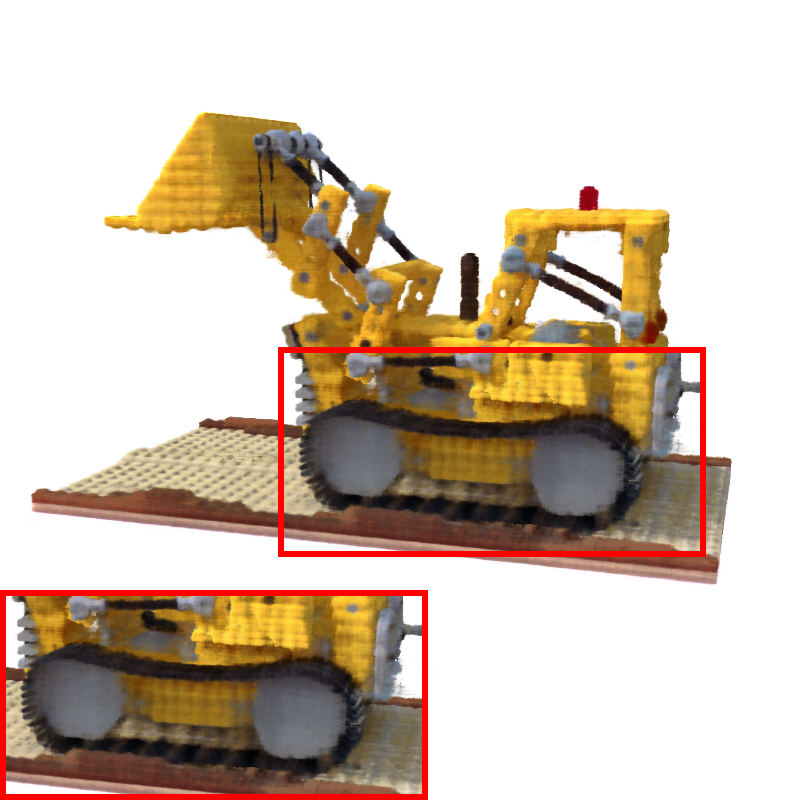}
         \caption{DietNeRF}
     \end{subfigure}
     \hfill
     \begin{subfigure}[b]{0.18\textwidth}
         \centering
         \includegraphics[width=\textwidth]{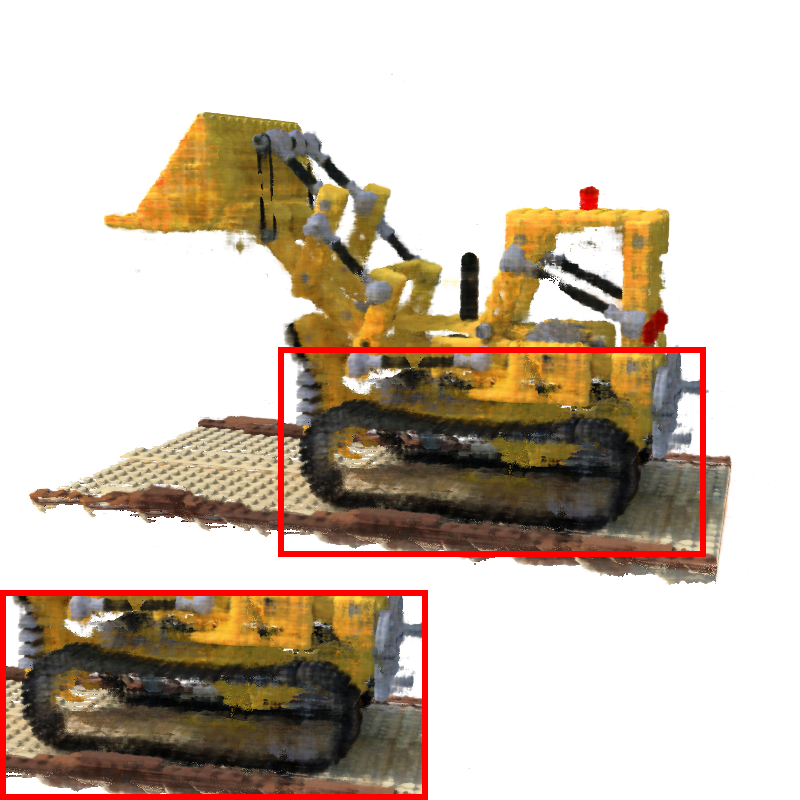}
         \caption{InfoNeRF}
     \end{subfigure}
     \hfill
     \begin{subfigure}[b]{0.18\textwidth}
         \centering
         \includegraphics[width=\textwidth]{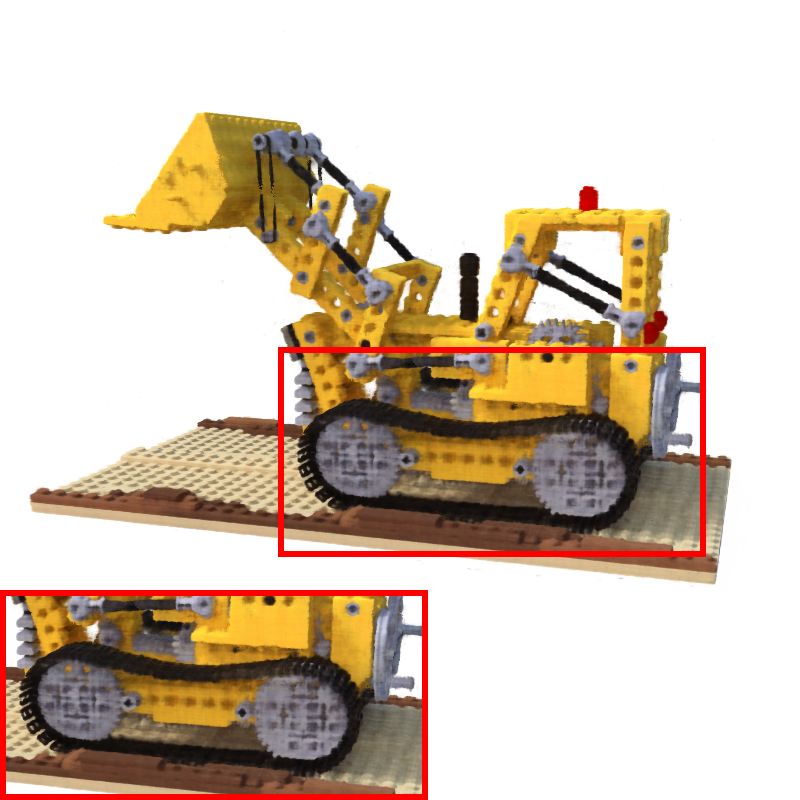}
         \caption{Ours}
     \end{subfigure}
     \hfill
     \begin{subfigure}[b]{0.18\textwidth}
         \centering
         \includegraphics[width=\textwidth]{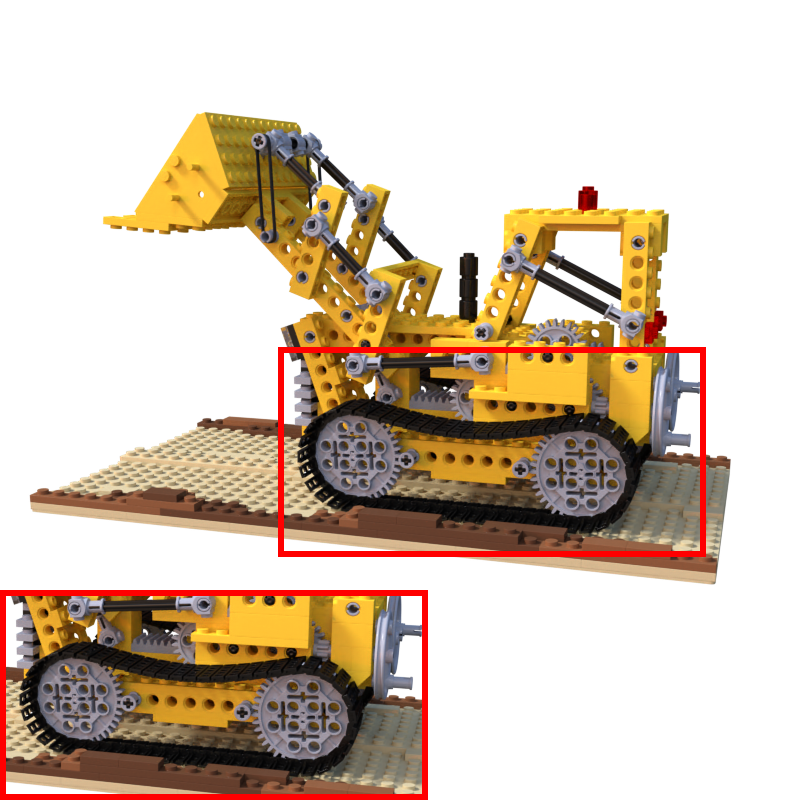}
         \caption{Ground Truth}
     \end{subfigure}
     \vspace{0.2cm}
    \caption{Qualitative results on the Blender dataset. Zoom in for a better view.}
    \label{fig:qual}
\end{figure*}
\begin{figure*}
     \centering
     \par\medskip
     \begin{subfigure}[b]{0.18\textwidth}
         \centering
         \includegraphics[width=\textwidth]{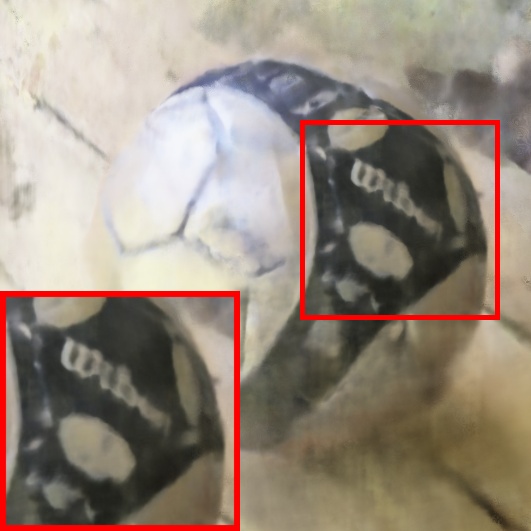}
     \end{subfigure}
     \hfill
     \begin{subfigure}[b]{0.18\textwidth}
         \centering
         \includegraphics[width=\textwidth]{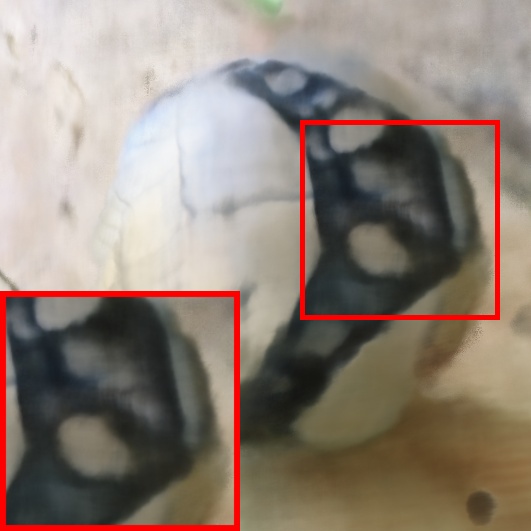}
     \end{subfigure}
     \hfill
     \begin{subfigure}[b]{0.18\textwidth}
         \centering
         \includegraphics[width=\textwidth]{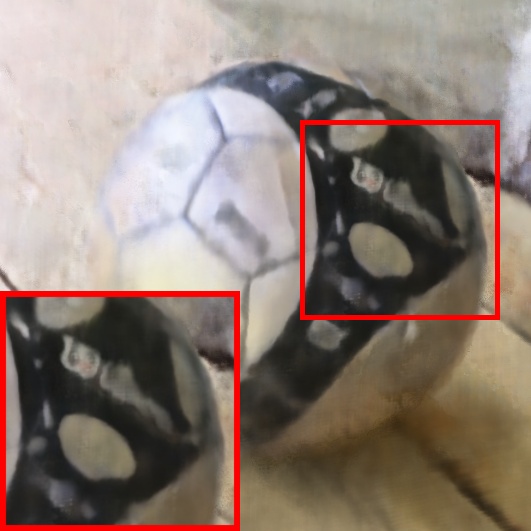}
     \end{subfigure}
     \hfill
     \begin{subfigure}[b]{0.18\textwidth}
         \centering
         \includegraphics[width=\textwidth]{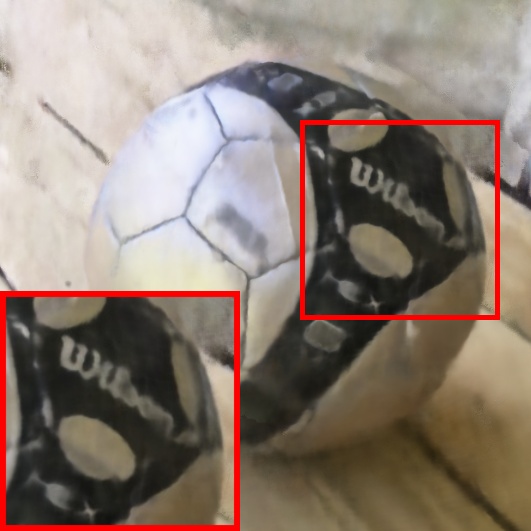}
     \end{subfigure}
     \hfill
     \begin{subfigure}[b]{0.18\textwidth}
         \centering
         \includegraphics[width=\textwidth]{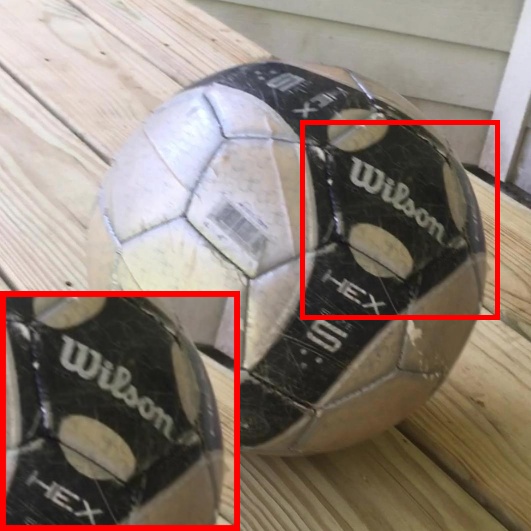}
     \end{subfigure}
     \par\medskip
     \begin{subfigure}[b]{0.18\textwidth}
         \centering
         \includegraphics[width=\textwidth]{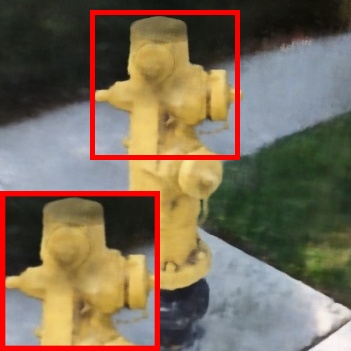}
     \end{subfigure}
     \hfill
     \begin{subfigure}[b]{0.18\textwidth}
         \centering
         \includegraphics[width=\textwidth]{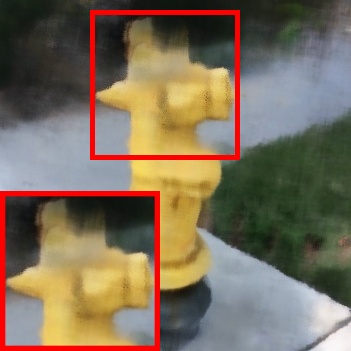}
     \end{subfigure}
     \hfill
     \begin{subfigure}[b]{0.18\textwidth}
         \centering
         \includegraphics[width=\textwidth]{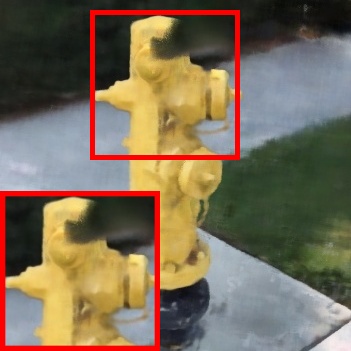}
     \end{subfigure}
     \hfill
     \begin{subfigure}[b]{0.18\textwidth}
         \centering
         \includegraphics[width=\textwidth]{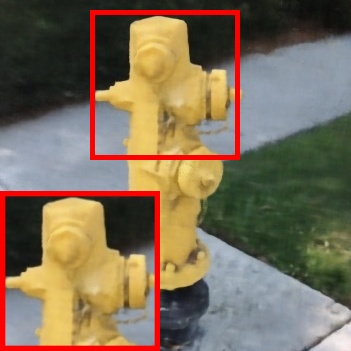}
     \end{subfigure}
     \hfill
     \begin{subfigure}[b]{0.18\textwidth}
         \centering
         \includegraphics[width=\textwidth]{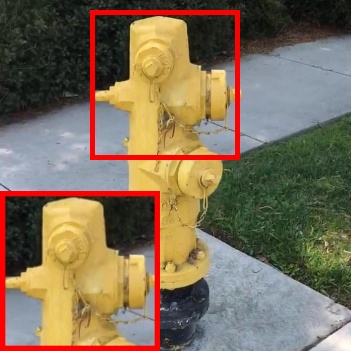}
     \end{subfigure}
     \par\medskip
     \begin{subfigure}[b]{0.18\textwidth}
         \centering
         \includegraphics[width=\textwidth]{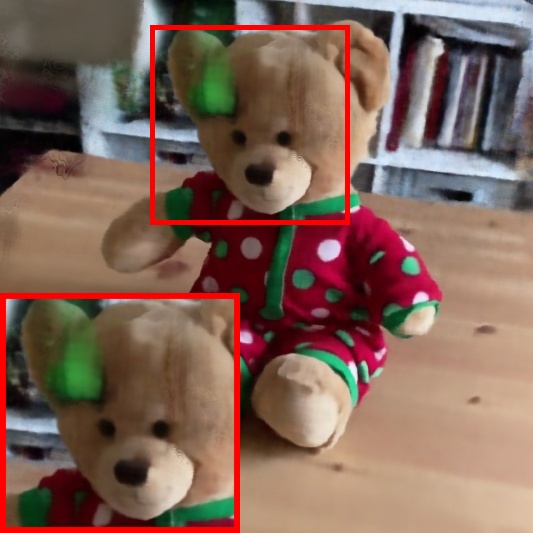}
         \caption{NeRF}
     \end{subfigure}
     \hfill
     \begin{subfigure}[b]{0.18\textwidth}
         \centering
         \includegraphics[width=\textwidth]{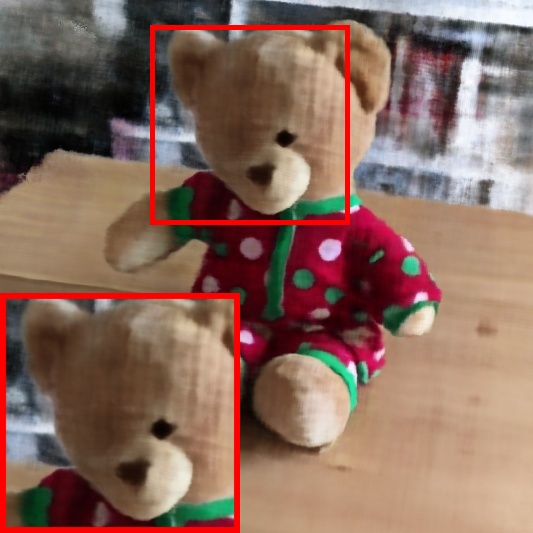}
         \caption{DietNeRF}
     \end{subfigure}
     \hfill
     \begin{subfigure}[b]{0.18\textwidth}
         \centering
         \includegraphics[width=\textwidth]{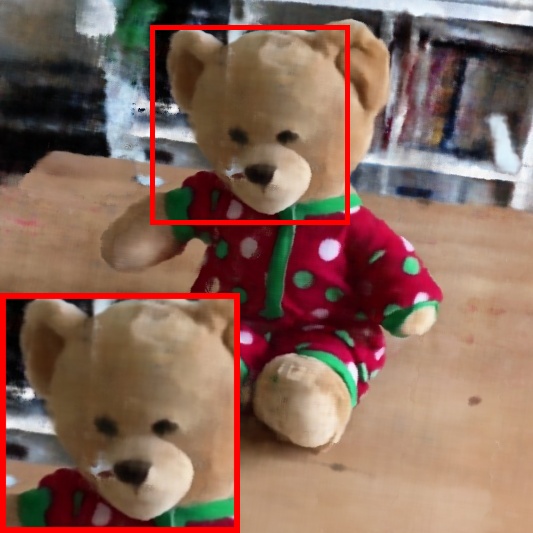}
         \caption{InfoNeRF}
     \end{subfigure}
     \hfill
     \begin{subfigure}[b]{0.18\textwidth}
         \centering
         \includegraphics[width=\textwidth]{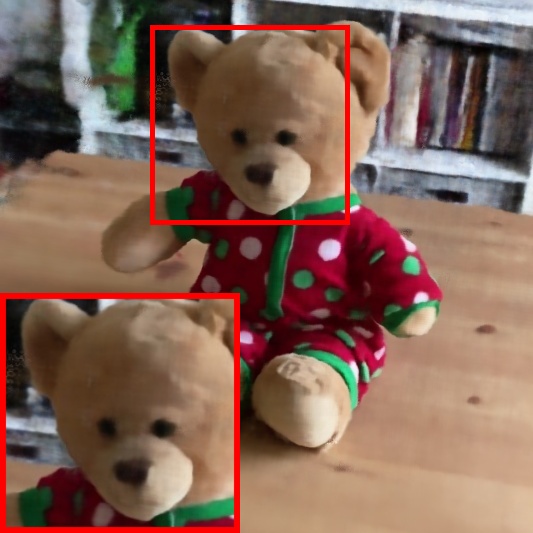}
         \caption{Ours}
     \end{subfigure}
     \hfill
     \begin{subfigure}[b]{0.18\textwidth}
         \centering
         \includegraphics[width=\textwidth]{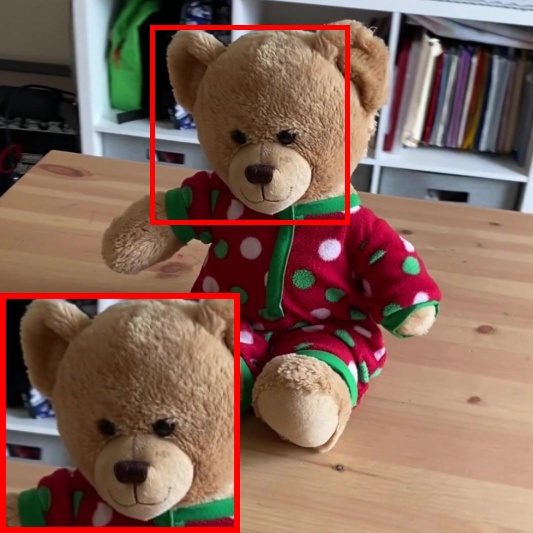}
         \caption{Ground Truth}
     \end{subfigure}
     \vspace{0.2cm}
    \caption{Qualitative results on the CO3D dataset. Zoom in for a better view.}
    \label{fig:qual_co3d}
\end{figure*}

\subsection{QUALITATIVE RESULTS}

The performance improvement of the proposed KeyNeRF in terms of rendering quality is visualized in Figure \ref{fig:qual} for the Blender dataset and in Figure \ref{fig:qual_co3d} for the CO3D dataset. The qualitative comparison against state-of-the-art methods shows that our informative rays selection strategy allows to render novel views with better details, especially in intricate structures such as the bulldozer wheels or the ship mast in Figure \ref{fig:qual}, and less hallucinated geometries (e.g. the teddybear and the hydrant in Figure \ref{fig:qual_co3d}). Moreover, our outputs are less blurry and preserve better the original colors of the scene. Finally, these results confirm that both DietNeRF \cite{dietnerf} and InfoNeRF \cite{infonerf} tend to saturate their improvements over the original NeRF \cite{nerf} in the considered setup, while our approach presents significant advantages. 

\section{\uppercase{Conclusion}}

In this work, we present KeyNeRF, a method to select informative samples for training a few-shot NeRF. Given a dense and redundant set of views, we first select a subset of cameras to ensure both scene coverage and baseline diversity. Then, individual rays for each view are sampled from a probability distribution based on the local entropy of the image. This two-stage process allows the optimization to focus on relevant information in early iterations, thus speeding up convergence with a limited training budget. In this way, KeyNeRF outperforms other few-shot methods on both synthetic and real-world data, without requiring additional inputs or auxiliary pre-trained networks. Moreover, our approach can be easily integrated into any NeRF codebase, as it only requires to change the lines of code for sampling the batch of rays to optimize at each iteration. The current formulation is limited by the object-centric acquisition trajectory and the assumption of the object having higher entropy than the background. As future work, we plan to tackle this issues and to integrate our selection procedure with other neural reconstruction methods.

\bibliographystyle{apalike}
{\small
\bibliography{paper}}

\begin{thebibliography}{}

\bibitem[Cao et~al., 2021]{3daug1}
Cao, M., Zheng, L., Jia, W., Lu, H., and Liu, X. (2021).
\newblock Accurate 3-d reconstruction under iot environments and its applications to augmented reality.
\newblock {\em IEEE Transactions on Industrial Informatics}, 17(3):2090--2100.

\bibitem[Chen et~al., 2022]{tensorf}
Chen, A., Xu, Z., Geiger, A., Yu, J., and Su, H. (2022).
\newblock Tensorf: Tensorial radiance fields.
\newblock In {\em European Conference on Computer Vision (ECCV)}.

\bibitem[Deng et~al., 2022]{dsnerf}
Deng, K., Liu, A., Zhu, J.-Y., and Ramanan, D. (2022).
\newblock Depth-supervised {NeRF}: Fewer views and faster training for free.
\newblock In {\em Proceedings of the IEEE/CVF Conference on Computer Vision and Pattern Recognition (CVPR)}.

\bibitem[Furukawa et~al., 2010]{furukawa}
Furukawa, Y., Curless, B., Seitz, S.~M., and Szeliski, R. (2010).
\newblock Towards internet-scale multi-view stereo.
\newblock In {\em 2010 IEEE computer society conference on computer vision and pattern recognition}, pages 1434--1441. IEEE.

\bibitem[Jain et~al., 2021]{dietnerf}
Jain, A., Tancik, M., and Abbeel, P. (2021).
\newblock Putting nerf on a diet: Semantically consistent few-shot view synthesis.
\newblock In {\em Proceedings of the IEEE/CVF International Conference on Computer Vision (ICCV)}, pages 5885--5894.

\bibitem[Kim et~al., 2022]{infonerf}
Kim, M., Seo, S., and Han, B. (2022).
\newblock Infonerf: Ray entropy minimization for few-shot neural volume rendering.
\newblock In {\em CVPR}.

\bibitem[Ladikos et~al., 2009]{ladikos}
Ladikos, A., Ilic, S., and Navab, N. (2009).
\newblock Spectral camera clustering.
\newblock In {\em 2009 IEEE 12th International Conference on Computer Vision Workshops, ICCV Workshops}, pages 2080--2086. IEEE.

\bibitem[Liu et~al., 2020]{nsvf}
Liu, L., Gu, J., Lin, K.~Z., Chua, T.-S., and Theobalt, C. (2020).
\newblock Neural sparse voxel fields.
\newblock {\em NeurIPS}.

\bibitem[Mauro et~al., 2014]{mauro}
Mauro, M., Riemenschneider, H., Signoroni, A., Leonardi, R., and Van~Gool, L. (2014).
\newblock An integer linear programming model for view selection on overlapping camera clusters.
\newblock In {\em 2014 2nd International Conference on 3D Vision}, volume~1, pages 464--471. IEEE.

\bibitem[Mildenhall et~al., 2020]{nerf}
Mildenhall, B., Srinivasan, P.~P., Tancik, M., Barron, J.~T., Ramamoorthi, R., and Ng, R. (2020).
\newblock Nerf: Representing scenes as neural radiance fields for view synthesis.
\newblock In {\em European conference on computer vision}, pages 405--421. Springer.

\bibitem[M{\"u}ller et~al., 2022]{instantngp}
M{\"u}ller, T., Evans, A., Schied, C., and Keller, A. (2022).
\newblock Instant neural graphics primitives with a multiresolution hash encoding.
\newblock {\em ACM Transactions on Graphics (ToG)}, 41(4):1--15.

\bibitem[Niemeyer et~al., 2022]{regnerf}
Niemeyer, M., Barron, J.~T., Mildenhall, B., Sajjadi, M. S.~M., Geiger, A., and Radwan, N. (2022).
\newblock Regnerf: Regularizing neural radiance fields for view synthesis from sparse inputs.
\newblock In {\em Proc. IEEE Conf. on Computer Vision and Pattern Recognition (CVPR)}.

\bibitem[Orsingher et~al., 2022a]{iciap}
Orsingher, M., Zani, P., Medici, P., and Bertozzi, M. (2022a).
\newblock Efficient view clustering and selection for city-scale 3d reconstruction.
\newblock In {\em Image Analysis and Processing--ICIAP 2022: 21st International Conference, Lecce, Italy, May 23--27, 2022, Proceedings, Part II}, pages 114--124. Springer.

\bibitem[Orsingher et~al., 2022b]{iv}
Orsingher, M., Zani, P., Medici, P., and Bertozzi, M. (2022b).
\newblock Revisiting patchmatch multi-view stereo for urban 3d reconstruction.
\newblock In {\em 2022 IEEE Intelligent Vehicles Symposium (IV)}, pages 190--196. IEEE.

\bibitem[Pan et~al., 2022]{activenerf}
Pan, X., Lai, Z., Song, S., and Huang, G. (2022).
\newblock Activenerf: Learning where to see with uncertainty estimation.
\newblock In {\em Computer Vision--ECCV 2022: 17th European Conference, Tel Aviv, Israel, October 23--27, 2022, Proceedings, Part XXXIII}, pages 230--246. Springer.

\bibitem[Perron and Furnon, 2022]{ortools}
Perron, L. and Furnon, V. (2022).
\newblock Or-tools.

\bibitem[Ramasinghe et~al., 2022]{frequencybias}
Ramasinghe, S., MacDonald, L.~E., and Lucey, S. (2022).
\newblock On the frequency-bias of coordinate-mlps.
\newblock In {\em Advances in Neural Information Processing Systems}.

\bibitem[Reizenstein et~al., 2021]{co3d}
Reizenstein, J., Shapovalov, R., Henzler, P., Sbordone, L., Labatut, P., and Novotny, D. (2021).
\newblock Common objects in 3d: Large-scale learning and evaluation of real-life 3d category reconstruction.
\newblock In {\em International Conference on Computer Vision}.

\bibitem[Roessle et~al., 2022]{densedepth}
Roessle, B., Barron, J.~T., Mildenhall, B., Srinivasan, P.~P., and Nie{\ss}ner, M. (2022).
\newblock Dense depth priors for neural radiance fields from sparse input views.
\newblock In {\em Proceedings of the IEEE/CVF Conference on Computer Vision and Pattern Recognition (CVPR)}.

\bibitem[Sch\"{o}nberger and Frahm, 2016]{sfm}
Sch\"{o}nberger, J.~L. and Frahm, J.-M. (2016).
\newblock Structure-from-motion revisited.
\newblock In {\em Conference on Computer Vision and Pattern Recognition (CVPR)}.

\bibitem[Seo et~al., 2023]{mixnerf}
Seo, S., Han, D., Chang, Y., and Kwak, N. (2023).
\newblock Mixnerf: Modeling a ray with mixture density for novel view synthesis from sparse inputs.
\newblock In {\em ArXiV}.

\bibitem[Sucar et~al., 2021]{imap}
Sucar, E., Liu, S., Ortiz, J., and Davison, A. (2021).
\newblock {iMAP}: Implicit mapping and positioning in real-time.
\newblock In {\em Proceedings of the International Conference on Computer Vision ({ICCV})}.

\bibitem[Van~der Merwe et~al., 2020]{3drob1}
Van~der Merwe, M., Lu, Q., Sundaralingam, B., Matak, M., and Hermans, T. (2020).
\newblock Learning continuous 3d reconstructions for geometrically aware grasping.
\newblock In {\em 2020 IEEE International Conference on Robotics and Automation (ICRA)}, pages 11516--11522.

\bibitem[Wang et~al., 2019]{3drob2}
Wang, Y., James, S., Stathopoulou, E.~K., Beltrán-González, C., Konishi, Y., and Del~Bue, A. (2019).
\newblock Autonomous 3-d reconstruction, mapping, and exploration of indoor environments with a robotic arm.
\newblock {\em IEEE Robotics and Automation Letters}, 4(4):3340--3347.

\bibitem[Wynn and Turmukhambetov, 2023]{diffusionerf}
Wynn, J. and Turmukhambetov, D. (2023).
\newblock Diffusionerf: Regularizing neural radiance fields with denoising diffusion models.
\newblock In {\em ArXiV}.

\bibitem[Yang et~al., 2023]{freenerf}
Yang, J., Pavone, M., and Wang, Y. (2023).
\newblock Freenerf: Improving few-shot neural rendering with free frequency regularization.
\newblock In {\em Proc. IEEE Conf. on Computer Vision and Pattern Recognition (CVPR)}.

\bibitem[Yeh and Lin, 2018]{3daug2}
Yeh, Y.-J. and Lin, H.-Y. (2018).
\newblock 3d reconstruction and visual slam of indoor scenes for augmented reality application.
\newblock In {\em 2018 IEEE 14th International Conference on Control and Automation (ICCA)}, pages 94--99.

\bibitem[Yen-Chen, 2020]{nerf-pytorch}
Yen-Chen, L. (2020).
\newblock Nerf-pytorch.
\newblock \url{https://github.com/yenchenlin/nerf-pytorch/}.

\end{thebibliography}

\end{document}